\documentclass[11pt]{article}

\usepackage{sectsty}
\usepackage{graphicx}

\usepackage[nopatch]{microtype}

\usepackage{longtable}
\usepackage{multirow}%
\usepackage{amsmath,amssymb,amsfonts}
\usepackage{algorithmic}
\usepackage{graphicx}
\usepackage{textcomp}
\usepackage{natbib}
\usepackage[inline]{enumitem}

\usepackage{bm}

\usepackage{units}
\usepackage{xeCJK}
\usepackage{fontspec}
\usepackage{xcolor}
\usepackage{linguex}
\usepackage[T1]{fontenc}
\usepackage[utf8]{inputenc}
\usepackage{booktabs}

\usepackage{xstring}
\usepackage{stfloats}

\title{Predicate-Argument Structure Divergences in Chinese and English Parallel Sentences and their Impact on Language Transfer}

\author{ Rocco Tripodi$^{1}$\thanks{Corresponding author: rocco.tripodi@unive.it}
  \and
  Xiaoyu Liu$^{2}$ \\
  $^{1}$Department of Environmental Sciences, Informatics and Statistics\\Ca' Foscari University of Venice, Dorsoduro 3246 - Venice, Italy \\
  $^{2}$Department of Linguistic Sciences And Foreign Literatures\\Catholic University of the Sacred Heart, Largo A. Gemelli, 1 - Milan, Italy
}

\begin{document}
\maketitle



\begin{abstract}
Cross-lingual Natural Language Processing (NLP) has gained significant traction in recent years, offering practical solutions in low-resource settings by transferring linguistic knowledge from resource-rich to low-resource languages.
This field leverages techniques like annotation projection and model transfer for language adaptation, supported by multilingual pre-trained language models. However, linguistic divergences hinder language transfer, especially among typologically distant languages.
In this paper, we present an analysis of predicate-argument structures in parallel Chinese and English sentences.
We explore the alignment and misalignment of predicate annotations, inspecting similarities and differences and proposing a categorization of structural divergences.
The analysis and the categorization are supported by a qualitative and quantitative analysis of the results of an annotation projection experiment, in which, in turn, one of the two languages has been used as source language to project annotations into the corresponding parallel sentences.
The results of this analysis show clearly that language transfer is asymmetric.
An aspect that requires attention when it comes to selecting the source language in transfer learning applications and that needs to be investigated before any scientific claim about cross-lingual NLP is proposed.
\end{abstract}

\section{Introduction}
Cross-lingual Natural Language Processing (NLP) \citep{agic-etal-2014-cross,ruder-etal-2019-transfer} has gained substantial momentum in recent years \citep{conneau-etal-2018-xnli,DBLP:conf/nips/ConneauL19}. At its core lies cross-lingual learning \citep{agic-etal-2014-cross,ponti-etal-2019-modeling}, a technique aimed at transferring linguistic knowledge from models trained on one language to those operating in another. This approach is particularly advantageous for low-resource languages, as it leverages data and models developed for resource-rich languages to overcome the scarcity of training materials.
A key method in this paradigm is annotation projection \citep{david2001inducing}, where annotations in one language are transferred to a parallel corpus in another.
The process involves identifying parallel sentences, aligning words, and assigning source-language labels to their corresponding target-language counterparts. Widely used in NLP tasks~\citep{pado2009cross,johannsen-etal-2016-joint,ni-etal-2017-weakly}, annotation projection has seen improved accuracy due to multilingual language models (mLMs) like mBERT \citep{devlin-etal-2019-bert} and XLM-R \citep{conneau-etal-2020-unsupervised}, which enhance word alignment~\citep{jalili-sabet-etal-2020-simalign,ijcai2021p539}.

The advent of mLMs has also revolutionized model transfer techniques, ushering in the \emph{pre-train, fine-tune, and predict} paradigm \citep{peters-etal-2019-tune}.
Under this framework, models are pre-trained on vast multilingual corpora using self-supervised objectives, fine-tuned for specific tasks in one language, and evaluated in \emph{zero-} or \emph{few-shot} settings on another language \citep{artetxe-schwenk-2019-massively}.
By leveraging multilingual knowledge during pre-training and specializing through fine-tuning, mLMs achieve strong cross-lingual performance \citep{pires-etal-2019-multilingual,conneau-etal-2020-unsupervised}.
However, recent research has challenged the efficacy of this paradigm, revealing significant \emph{data contamination} in multilingual corpora due to misidentified languages \citep{blevins-zettlemoyer-2022-language}.
This contamination introduces noisy or erroneous data during training, undermining the acquisition of multilingual knowledge and limiting the reliability of cross-lingual transfer.

Despite advancements, cross-lingual divergences remain a major obstacle to the efficacy of both annotation projection and model transfer techniques \citep{gerz-etal-2018-relation,DBLP:journals/ipm/EronenPM23}.
These divergences reduce the effectiveness of parallel corpora \citep{carpuat-etal-2017-detecting,vyas-etal-2018-identifying} and negatively impact cross-lingual model performance in tasks such as machine translation \citep{dorr1994machine,vyas-etal-2018-identifying} and Abstract Meaning Representation (AMR) parsing \citep{blloshmi-etal-2020-xl,wein-schneider-2021-classifying}.
Identifying and analyzing linguistic divergences is crucial for understanding the limits of cross-lingual knowledge transfer \citep{ruder-etal-2019-transfer} and linguistic structure generalization \citep{yao-koller-2022-structural}.
However, as we will show, available resources for studying linguistic divergences like the World Atlas of Language Structures (WALS) \citep{wals}, lack of direct applicability to task-specific NLP tasks such as Semantic Role Labeling (SRL) and AMR.

Studying linguistic divergences not only illuminates the constraints of multilingual language modeling \citep{gerz-etal-2018-relation} but also informs the creation of more suitable multilingual annotation guidelines \citep{xue-etal-2014-interlingua,nikolaev-etal-2020-fine}.
However, this endeavor is inherently complex due to the enormous parameter space of mLMs and the intricate nature of cross-lingual divergences.
Research shows that cross-lingual model performance declines when target languages are typologically distant from the source \citep{lauscher-etal-2020-zero,ponti-etal-2021-parameter}.
Yet, there has been little empirical work systematically quantifying the prevalence and types of divergences driving this trend.
This gap leads to two significant issues.
First, English is often the default source language for cross-lingual transfer \citep{DBLP:journals/ipm/EronenPM23}, potentially biasing models toward Indo-European languages, particularly Germanic and Romance families \citep{agic-etal-2016-multilingual}.
Second, critical divergences between typologically distant languages are overlooked, limiting the ability of mLMs to generalize structurally \citep{yao-koller-2022-structural}.
Highlighting these structural issues is essential for mitigating biases, improving cross-lingual NLP applications, and advancing the development of truly multilingual AI systems.

For this reason, we decided to investigate a specific type of divergences that occur at the semantic level and in particular in the predicate-argument structure of professionally translated Chinese and English parallel sentences.
{We selected English and Chinese for this study due to their typological distance, as they belong to vastly different language families (Germanic and Sino-Tibetan, respectively).
This significant linguistic divergence makes them an ideal pair for examining structural and semantic discrepancies, which are critical for understanding the challenges of cross-lingual transfer in NLP.
Additionally, English and Chinese are among the languages with the most extensive linguistic resources, providing a robust foundation for analysis.
For this reason, they are extensively used as source languages in transfer learning applications.
However, it is important to note that for less-resourced languages, the situation could be even more challenging.
The insights gained from studying this typologically distant and resource-rich language pair can thus serve as a baseline to inform future research on lower-resource languages, where structural divergences may further exacerbate the difficulties of cross-lingual NLP.}

We define predicate-argument structure divergences as cases in bilingual sentence pairs where different predicates and/or arguments are used to convey the same meaning.
To identify such misalignments, we conducted a manual analysis of a parallel corpus annotated with predicate-argument structures and verb sense information.
Our methodology involved two main stages.
First, we performed a quantitative analysis comparing the distribution of verbs and semantic annotations in the reference corpus, examining variations in the number of verbs across parallel sentences and their effect on semantic structures. Second, we conducted an in-depth qualitative analysis of alignment and misalignment cases, identifying four distinct types of divergences.
These were characterized, quantified, and exemplified to provide insights into their nature and causes.
The goal of this work is to understand the occurrence of predicate-argument misalignments, characterize and quantify them, and explain why even high-quality translations often introduce meaning mismatches. For instance:

\ex.[]
\a. John broke into the room.
\b. Juan forzó la entrada al cuarto.

\exg.[] Juan forzó la entrada al cuarto\\
John forced the entry {to the} room\\

In which, \emph{broke into} is translated using a completely different expression, i.e., \emph{forced the entry}.
Translations, in fact, can introduce divergences at different levels: lexical choices between near-synonyms, for example, can produce language-specific nuances \citep{edmonds-hirst-2002-near}; typological divergences can introduce structural mismatches \citep{dorr1994machine}; furthermore, the production of a divergent translation by a professional translator may be just the result of the fact that it sounds more natural and fluent in the target language.

The contribution of this article can be summarized as follows:
\begin{itemize}
    \item An in-depth analysis of predicate-argument structure divergences in Chinese and English;
    \item A linguistically motivated classification of predicate-argument structure divergences;
    \item A manually annotated parallel corpus for testing SRL annotation projection models;
    \item An evaluation of SRL annotation projection, that highlights language transfer asymmetry;
    \item An explanation of cross-lingual transfer ineffectiveness in SRL.
\end{itemize}

The structure of the paper is as follows: Section 2 provides an overview of linguistic divergence studies in the NLP field. Section 3 introduces key concepts foundational to the analysis in Section 4, which focuses on a quantitative and qualitative examination of predicate-argument structure divergences. Section 5 compares human annotations with automatic annotation projection for the SRL task. Finally, Section 6 concludes the paper and outlines potential directions for future research.



\section{Related Work}
The analysis of linguistic divergences is a critical topic in NLP, explored across various subfields, including part-of-speech tagging \citep{agic-etal-2016-multilingual}, dependency parsing \citep{mcdonald-etal-2013-universal}, cross-lingual word embeddings \citep{ormazabal-etal-2019-analyzing}, semantic parsing \citep{xue-etal-2014-interlingua,wein-etal-2022-effect}, and machine translation \citep{vyas-etal-2018-identifying}, among others.

Drawing from existing definitions of linguistic divergence, we identified two primary research approaches.
The first focuses on semantic divergences in pseudo-parallel data, such as automatically aligned sentences from multilingual resources like Wikipedia articles \citep{schwenk-etal-2021-wikimatrix} or machine-translated corpora.
These studies typically aim to detect divergent sentence pairs that could introduce noise into models trained for tasks such as machine translation \citep{vyas-etal-2018-identifying,carpuat-etal-2017-detecting,briakou-carpuat-2020-detecting}.

The second line of work examines broader linguistic divergences—encompassing morphology, syntax, and semantics—within professionally translated parallel corpora that aim to convey equivalent meaning.
Our study aligns with this second approach, as it investigates general linguistic divergences in translations.
Even high-quality translations often exhibit structural mismatches that affect syntax and semantics, challenging the assumption of semantic equivalence.

A foundational study on translation divergences by \citep{dorr1994machine} proposed six categories abstracted from syntactic details to highlight general linguistic divergences:
\begin{itemize}
    \item Thematic: A theme in the source language (SL) is realized as the verbal object in the target language (TL).
    \item Promotional: A modifier in the SL is promoted to a main verb in the TL.
    \item Demotional: The opposite of promotional, where a main verb in the SL becomes a modifier in the TL.
    \item Structural: Argument realization differs syntactically between languages.
    \item Conflational: Two or more words in one language are translated into a single word in another language.
    \item Categorial: Words change their part-of-speech (POS) tags during translation.
\end{itemize}
While \citep{dorr1994machine} offered linguistically motivated examples, they did not quantify the distribution of these categories, leaving the significance of the problem unmeasured.

Building on these ideas, \citep{nikolaev-etal-2020-fine} analyzed cross-linguistic morphosyntactic divergences (CLMD) in Universal Dependencies \citep[UD]{nivre-etal-2016-universal} parallel corpora for English, French, German, Italian, and Spanish.
Using a language-independent framework, they identified divergence patterns by comparing dependency paths between aligned content words.
This analysis revealed substantial misalignment in syntactic relations and suggested that refining UD guidelines and normalizing cross-linguistic constructions could reduce divergences. \citep{arviv-etal-2021-relation} demonstrated the efficacy of such modifications in a cross-lingual zero-shot setting.
Morphosyntactic divergences are more apparent than semantic ones, which are less intuitive.
Despite the expectation that parallel sentences should convey identical meanings, semantic differences often arise.

The possibility of creating a universal semantic representation for parallel sentences—similar to UD for syntax—has attracted considerable attention, as such a framework could be transformative for multilingual applications like machine translation and dialogue systems.
Semantic parsing frameworks such as Abstract Meaning Representation \citep[AMR]{Banarescu2013} provide a formalism for representing meaning as rooted, directional, labeled graphs encoding predicate-argument structures along with temporal and locative information. \citep{xue-etal-2014-interlingua} examined AMR’s potential as a transfer layer for machine translation, analyzing 100 manually annotated English-Chinese and English-Czech sentence pairs. While AMR structures aligned well in many cases, non-local structural divergences were observed, particularly between English and Czech, where 53
In another study, \citep{sulem-etal-2015-conceptual} compared English and French translations using Universal Conceptual Cognitive Annotation \citep[UCCA]{abend-rappoport-2013-universal}, another semantic representation formalism. UCCA graphs, which represent meaning as directed acyclic graphs (DAGs), demonstrated greater stability across English-French translations than AMR graphs, corroborating other findings on compatibility between these typologically similar languages \citep{pires-etal-2019-multilingual,keung-etal-2020-dont,blloshmi-etal-2020-xl}. However, a systematic explanation for this phenomenon is still lacking.
Lastly, \citep{vsindlerova2013verb} analyzed predicate-argument structure divergences in the Prague Czech-English Dependency Treebank \citep{hajic-etal-2012-announcing}. Focusing on judgment verbs annotated with PDT-VALLEX \citep{hajic2003pdt}, the study highlighted significant divergences in the alignment of argument roles, particularly between the \textsc{addressee} and \textsc{patient} roles.

{A linguistic resource that offers a foundation for computing linguistic divergences is The World Atlas of Language Structures (WALS) \citep{wals}.
It consists of a database cataloging phonological, grammatical, and lexical properties for over 2,600 languages.
Despite this wide coverage, comparisons across languages using WALS are challenging due to its sparse feature representation and lack of direct applicability to task-specific NLP datasets like Semantic Role Labeling (SRL) and AMR.
As demonstrated in our study, only one of the divergences observed could be detected with WALS, underscoring its limitations for fine-grained NLP analysis.}


\begin{table*}[!h]
\caption{Example of VerbAtlas frames and arguments structures associated with the verb \emph{say}}\label{tab:VA}
\resizebox{1\textwidth}{!}{
\begin{tabular}{p{2.5cm} p{2cm} p{4cm} p{9cm} }
\toprule
\textbf{Frame} & \textbf{BN ID} &\textbf{Description} & \textbf{Argument structure}\\ 
SPEAK & bn:00093287v & Express in words & An agent SPEAKS about a topic in a location to a recipient using an instrument achieving a result (+attribute)\\
SPEAK & bn:00093290v & Utter aloud & An agent SPEAKS about a topic in a location to a recipient using an instrument achieving a result (+attribute)\\
DECREE DECLARE & bn:00093291v & State as one's opinion or judgement; declare & An agent DECREES-DECLARES result a theme regarding a topic to a recipient (+attribute)\\
AFFIRM & bn:00082527v & Report or maintain & An agent AFFIRMS a theme to a recipient (+attribute) \\
CITE & bn:00092425v & Have or contain a certain wording or form & An agent CITES a theme from a source (+attribute)\\
\bottomrule
\end{tabular}}
\end{table*}

\section{Methodology}
In this study, we randomly sampled 400 parallel sentences in English and Chinese from the \textsc{UniteD-SRL} dataset \citep{tripodi2021united}, a new benchmark for Semantic Role Labeling (SRL) and Verb Disambiguation (VD) encompassing predicate-argument structure annotations in four languages, namely Chinese, English, French and Spanish.
The sentences in this dataset have been extracted from the UN Parallel Corpus \citep{DBLP:conf/lrec/ZiemskiJP16}, a multilingual compilation of official United Nations records and parliamentary reports. This corpus includes more than 11 million sentences per language, spanning 86,000 documents and categorized into 18 semantic domains.
In UniteD-SRL, verbs are marked with two layers of semantic annotation, one is fine-grained and the other one is coarse-grained.
Fine-grained information is obtained from BabelNet~\citep{navigli2012babelnet} an innovative multilingual encyclopaedic dictionary whose structure is similar to WordNet \citep{miller-1992-wordnet} and is based on the notion of synsets, a group a words that refer to the same concept.
However, in BabelNet, the notion of synset is extended to include multilingual lexicalizations for the same concept.
Each concept in this resource is associated to an unique identifier (BabelNet ID).
The coarse-grained information is provided at the frame level using VerbAtlas \citep{di2019verbatlas}, a novel and comprehensive semantic resource for verbs that in contrast to other popular lexical resources of this kind, such as PropBank \citep{Palmer2005} or FrameNet \citep{baker-etal-1998-berkeley-framenet}, adopts 25 explicit semantic roles shared across all its frames.
The VerbAtlas definitions of the frames used in this work are reported in Table \ref{tab:frame_definitions} and the defintion of the semantic roles are shown in Table \ref{tab:roles_definitions} (both in Appendix \ref{sec:appendix}).
As an example, in Table \ref{tab:VA}, we can find some of the frames associated with the verb \emph{say}.
From this table, it is possible to see that even if a verb can be associated to the same frame, e.g., \textsc{SPEAK} in our example, the BabelNet ID indicates a specific sense of the frame, e.g., the two frames \textsc{SPEAK} associated to two different IDs, defined \textit{Express in words} and \textit{Utter aloud} respectively.
Furthermore, each frame is accompanied by a prototypical argument structure, e.g., the first frame, \textsc{SPEAK}, includes six semantic roles: \textit{agent}, \textit{topic}, \textit{location}, \textit{recipient}, \textit{instrument} and \textit{result}.
As we can see from the example, the same frames share the same argument structure, despite the different senses, while different frames have distinct argument structures.
Thanks to the connection to BabelNet synsets, the information in VerbAtlas can be considered multilingual, the verb \emph{speak} in English and the verb \emph{hablar} in Spanish could be annotated with the same sense, frame and argument structure.
This allowed UniteD-SRL to develop parallel annotations for multilingual span- and dependency-based SRL that can be easily compared.
This dataset is sourced from the United Nations Parallel Corpus \citep{ziemski-etal-2016-united} and includes official UN documents and parliamentary reports from ten semantic domains.



{The annotations of the considered sentences were manually compared in both directions: using Chinese as the source language (SL) and English as the target language (TL), as well as vice versa. The comparisons were primarily aimed at identifying predicate divergences that automatically propagate through the corresponding argument structures. For each pair of sentences, the methodology followed these steps: \begin{enumerate}[label=(\roman*)] \item identification of predicates; \item alignment of predicates; and \item analysis of the resulting alignments/misalignments.
\end{enumerate}
The verbs considered in this study specifically exclude modal and light verbs.
To ensure the reliability of the analysis, two independent annotators performed the manual alignment and misalignment assessments, following the same set of guidelines.
At the end of the annotation, problematic cases were discussed and a unique annotation was provided.
This sanity check underscores the robustness of the manual analysis, allowing for a more confident presentation of both quantitative and qualitative results.}

\subsection{Example}
In this section we present an example of analysis in which we show two parallel sentences in English (EN) and Chinese (ZH) and provide an English proto-translation of the Chinese sentences.
In the proto-translation, each Chinese word is associated with its literal translation, that can be found under the corresponding logograms.
Frame alignments are indicated in green, while frame misalignments are indicated in red.
The following and all the other examples in this article are numbered and consist of three parts: 2 parallel sentences (a. and b.) followed by the English proto-translation of the Chinese sentence.

\ex.
\a. Mr. Repasch (USA) \textcolor{green!50!.}{said} that he would \textcolor{red!50!.}{prefer} to \textcolor{green!50!.}{continue} discussion of those matters in informal consultations. 
\b. Repasch 先生(美国)\textcolor{green!50!.}{说}，他 \textcolor{red!50!.}{希望}在非正式协商时\textcolor{green!50!.}{继续}\textcolor{red!50!.}{讨论}这个问题。

\exg.[] Repasch 先生 (美国) 说, 他 希望 在 非 正式 协商 时 继续 讨论 这个 问题\\
    Repasch Mr. USA say, he hope in non formal consultation time continue discuss this matter\\

\noindent These sentences encompass different predicate-argument structures.
In English there are three verbs, we list their frames and the semantic roles of the arguments separately in the examples below:
\begin{enumerate}
\item Mr. Repasch\textsubscript{agent} (USA) \textcolor{green!50!.}{said}\textsubscript{verb (frame: AFFIRM)} that he would prefer to continue discussion of those matters in informal consultations\textsubscript{theme}.
\item Mr. Repasch (USA) said that he\textsubscript{agent} would \textcolor{red!50!.}{prefer}\textsubscript{verb (frame: CHOOSE)} to continue discussion of those matters in informal consultations\textsubscript{theme}. 
\item  Mr. Repasch (USA) said that he\textsubscript{agent} 
would prefer to \textcolor{green!50!.}{continue}\textsubscript{verb (frame: CONTINUE)}
discussion of those matters in informal consultations\textsubscript{theme}.
\end{enumerate}

\noindent The corresponding Chinese sentence contains four verbs:

\begin{enumerate}
\item Repasch 先生(美国) \textsubscript{agent} \textcolor{green!50!.}{说}\textsubscript{verb (frame: AFFIRM)}，他希望在非正式协商时继续讨论这个问题。\textsubscript{theme}
\item 他 \textsubscript{agent} \textcolor{red!50!.}{希望} \textsubscript{verb (frame: REQUIRE\_NEED\_WANT\_HOPE)} 在非正式协商时\textsubscript{source} 继续讨论这个问题\textsubscript{theme}
\item 他\textsubscript{agent}希望在非正式协商时\textcolor{green!50!.}{继续} \textsubscript{verb (frame: CONTINUE)} 讨论这个问题\textsubscript{theme}
\item 他\textsubscript{agent}
希望在非正式协商时继续\textcolor{red!50!.}{讨论} \textsubscript{verb (frame: DISCUSS)} 这个问题\textsubscript{topic}
\end{enumerate}


\begin{figure*}[!ht]
\begin{center}
\includegraphics[width=.99\textwidth,trim={0cm 0cm 6.05cm 0cm},clip]{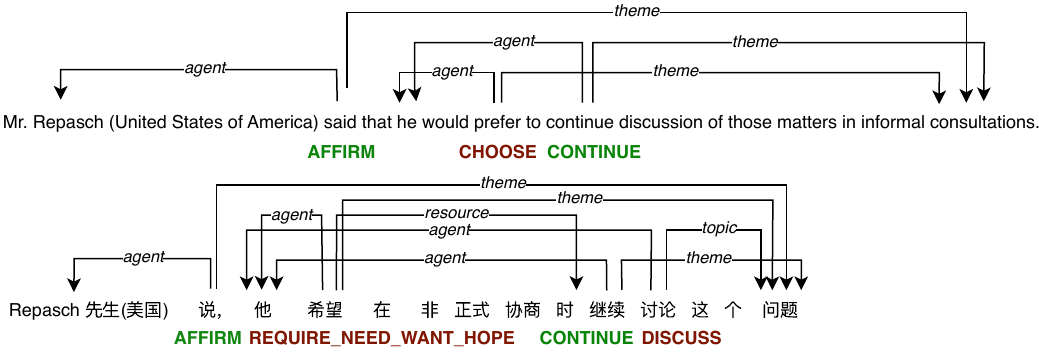} 
\caption{Example of predicate-argument divergence. Aligned frames are in green and disaligned frames are in red.}\label{fig:example}
\end{center}
\end{figure*}

\noindent In the example above, also presented in graphical form in Figure \ref{fig:example}, we can observe that the verbs \emph{said}/\emph{说} and \emph{continue}/\emph{继续} share the same frame annotation with consistent semantic roles for their arguments.
However, when it comes to the verb \emph{prefer}, it is translated in Chinese with a different verb, i.e., \emph{希望}/\emph{hope}, that consequently requires different frame annotations.
In this context, the semantic role of the argument \emph{in informal consultations}/\emph{在非正式协商时} changes from being the \textit{theme} (of the verb \emph{prefer}) to the \emph{source} (of the verb \emph{希望}/\emph{hope}).
Furthermore, there is an additional verb in Chinese, \emph{讨论}/\emph{discuss}, which includes a new role (\textit{topic}) to the argument \emph{这个问题}/\emph{in informal consultations}.

The proposed example allows to describe different cases of alignment and misalignment.
It shows how professionally translated sentences can introduce meaning mismatches, making parallel sentences not perfectly equivalent.

\begin{figure}[!ht]
\includegraphics[width=0.5\linewidth,trim={.1cm .1cm .5cm .5cm},clip]{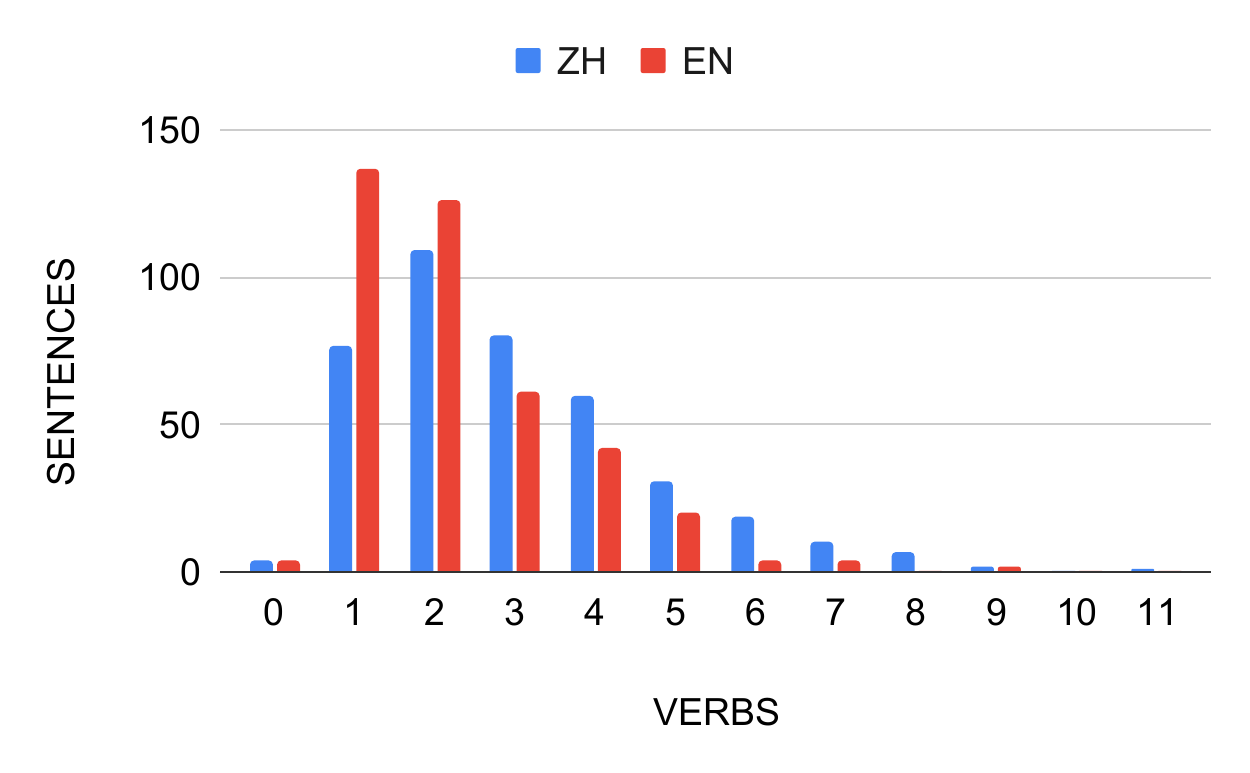}
\caption{Predicates distribution in both languages}\label{fig:preddist}
\end{figure}

\section{Results and Analysis}
The qualitative and quantitative results of our analyses are presented in this section.
In Section \ref{sec:pred_dist} the distribution of frames in the two languages studied is presented and complemented with the analysis of language-specific frames in Sextion \ref{sec:lang_spec_frames}.
A categorization of alignments and misalignments cases is then presented in Section \ref{sec:categorization}.

\subsection{Predicate distributions}\label{sec:pred_dist}
The first analysis that we conducted is concerned with the distribution of predicate-argument structures in the two languages of our corpus.
This analysis allows us to see that in {194} parallel sentences ({$48\%$}), the number of Chinese verbs exceeded that of English verbs.
In {177} sentences ({$45\%$}), the number of verbs was the same in both languages, while in {29} sentences ($7\%$), the number of Chinese verbs was lower than that of English verbs.
To better understand how predicates are distributed among sentences, we then computed the number of sentences that contain a defined number of verbs in them.
The results of this computation are reported in Figure \ref{fig:preddist}.
As shown in the plot, the most common number of verbs in an English sentence is one or two, accounting for {$66\%$} of all sentences.
In contrast, Chinese sentences exhibit more variation, with verb counts ranging from one to five, and a notable concentration of sentences containing one to three verbs.
The main observation that emerges from these distributions is that in Chinese sentences, the number of predicates is much higher.
In fact, in Chinese sentences we can find {$1198$} predicates, while in English there are {$909$}.
This simple analysis shows clearly that at least {$24\%$} of the verbs in Chinese could not be aligned to parallel sentences in English.
Analyzing the sentences that diverge for the number of predicates, we found that this discrepancy is contingent upon the different use of stative and dynamic verbs in the two languages.
In English, verbs are used mainly in stative expressions, whereas in Chinese verbs are predominant and are used for both stative and dynamic expressions.
A direct consequence of the higher number of verbs in Chinese sentences, as we will present in detail in Section \ref{sec:categorization}, is the use of nominalization of verbal expression in English, i.e., an action verb in a Chinese sentence can be aligned with a nominal phrase in the English corresponding sentence.
In fact, English often relies on a single predicate to convey the primary action.
Additional actions are expressed using infinitive forms, continuous tense, gerunds, adjectives, or other constructs with action-related meanings.
In contrast, Chinese tends to employ a multi-verb construction when encountering additional actions, achieving a more dynamic effect.
An example of this phenomenon is presented below.

\ex.
\a. I have the honour to forward herewith the text of a declaration entitled: "Declaration on \textcolor{green!50!.}{Promoting} the universalization of the Convention on the \textcolor{red!50!.}{Prohibition} of the \textcolor{red!50!.}{Use, Stockpiling, Production} and \textcolor{red!50!.}{Transfer} of Anti-Personnel Mines and on Their \textcolor{red!50!.}{Destruction}" that was adopted ...
\b. 我谨随信转交...通过的一项宣言的全文，题为"关于\textcolor{green!50!.}{促进}各国普遍\textcolor{red!50!.}{加入}《关于\textcolor{red!50!.}{禁止使用、储存、生产}和\textcolor{red!50!.}{转让}杀伤人员地雷及\textcolor{red!50!.}{销毁}此种地雷的公约》的宣言"。

\exg.[] 我 谨 随 信 转交 ... 通过 的 一项 宣言 的 全文， 题 为 "关于 促进 各 国 普遍 加入 《关于 禁止 使用、 储存、 生产 和 转让 杀伤 人员 地雷 及 销毁 此 种 地雷 的 公约》 的 宣言"。\\ 
   I sincerely with letter transfer ... pass of one declaration of fullpage title is "about promote every country universalise join "about prohibit use, store, produce and transfer kill person mine and destroy this type mine of convention" of declaration.    \\

\noindent In this example, we can see that \emph{promoting} is aligned with \emph{促进}/\emph{promote}, and that \emph{禁止}/\emph{prohibit},\emph{使用}/\emph{use},\emph{储存}/\emph{store}, \emph{生产}/\emph{produce},\emph{转让}/\emph{transfer},\emph{销毁}/\emph{destroy} are using nominalization of verbal expression in English.
Furthermore, the verb \emph{加入}/\emph{join} is not used in English.

\subsection{Language-specific frames}\label{sec:lang_spec_frames}
Our study on predicate divergences is complemented by the analysis of the frames associated with each verb.
The first observation that we can make is that, among the {$177$} sentences with an equal number of verbs, only {$52$} ({$13\%$} of our sample) exhibited identical annotation both at the frame and at the sense level.
This result makes clear how difficult it would be to use annotation projection techniques on this dataset since only in a few cases it would be possible to transfer a full annotation from one language to another.

We also conducted a frame analysis of verbs in all sentences to see how frames are distributed in both languages.
The results revealed that the frames for these verbs are not entirely shared.
There are 39 frames associated with English verbs that do not appear in Chinese, for example, \textsc{regret\_sorry} and \textsc{subjugate}.
While 41 frames associated with Chinese verbs are absent in English, for example, \textsc{border} and \textsc{generate}.
The most common shared frames include: \textsc{include-as} that appears 37 times in ZH ($4\%$) and 23 times in EN ($3\%$); \textsc{carry-out-action}, appearing 49 times in ZH ($5\%$) and 23 times in EN ($3\%$); \textsc{propose}, 28 times in ZH ($3\%$) and 20 times in EN ($3\%$): and \textsc{increase\_enlarge\_multiply}, 28 times in ZH ($3\%$) and 21 times in EN ($3\%$).
From these results emerges that even common frames are not always aligned.

\begin{table}[h]
\caption{Number of verbs in each category.}\label{tab:divcat}
\centering
\begin{tabular}{lll}
\toprule
\textbf{cat.} & \textbf{ZH as SL} & \textbf{EN as SL}\\
 1 & {$537$ (44.8\%)} & {$537$ (59.1\%)} \\
 2 & {$240$ (20.0\%)} & {$244$ (26.8\%)} \\
 3 & {$284$ (23.7\%)}  & {$76$ (8.3\%)} \\
 4 & {$137$ (11.4\%)} & {$52$ (5.7\%)} \\
\bottomrule
\end{tabular}
\end{table}

\subsection{Categorization of divergences}\label{sec:categorization}
The analysis of aligned predicates led to the identification of 4 main cases:
\begin{enumerate}
\item Frame convergence: it is possible to align verbs in parallel sentences and to find corresponding verbs annotated with the same frame and sense;
\item Frame divergence: it is possible to align verbs in parallel sentences but corresponding verbs are annotated with different frames and/or senses;
\item Non-verbal alignment: it is not possible to align verbs in parallel sentences, however, it is possible to align verbs in one language with non-verb forms in the other language;
\item Misalignment: it is not possible to find a corresponding verb translation in the target language.
\end{enumerate}

\noindent We provide the distribution of each category in Table \ref{tab:divcat}.
From these statistics, we can see that despite only a few sentences being completely aligned in our corpus (see \S \ref{sec:pred_dist}), the percentage of single frames that can be aligned is higher and corresponds to {$59.1$} if we use English as SL and {$44.8$} if we use Chinese as SL.
When English is used as the SL, most instances of verb misalignment stem from differences in frame annotation.
Furthermore, we can also observe that when dealing with the third and fourth categories, a significant portion of Chinese verbs is translated into non-verb forms or left untranslated when Chinese is used as the SL.
However, this occurrence is highly reduced when English is the source language.
The analyses conducted on each category are presented in detail in a dedicated section, in which, we identified frequently occurring verbs within each category and retraced them to their occurrences in parallel sentences for in-depth analysis.


\subsubsection{Frame convergence}
We included predicate pairs in the frame convergence category if they share the same VerbAtlas frame and BabelNet ID.
As an example, we can consider the following parallel sentences:
\ex.
\a. In El Salvador, women's participation in higher education \textcolor{green!50!.}{increased} from 9.6\% in 1995 to 15.9\% in 2002.
\b. 在萨尔瓦多，妇女接受高等教育的比例从 1995 年的 9.6\%\textcolor{green!50!.}{上升}到 2002 年的 15.9\%。

\exg.[] 在 萨尔瓦多， 妇女 接受 高等 教育 的 比例 从 1995 年 的 9.6\% 上升 到 2002 年 的 15.9\%。\\
    In Salvador, women accept high-level education of rate from 1995 year of 9.6\% increase to 2002 year of 15.9\%\\

\noindent Both the verb \emph{increase} and \emph{上升}/\emph{increase} have been annotated with the same VerbAtlas frame, i.e.,  \textsc{increase\_enlarge\_multiply}, and associated with the BabelNet synset bn:00085125v described as \emph{Increase in value or to a higher point}.
These cases of perfect lexical, syntactic, and semantic alignment are categorized as frame convergence and result in the most frequent category even though they account for less than 60\% of the verbs in both languages.

The most common aligned frames in both languages are: \textsc{carry-out-action} that occurs 49 times in Chinese is used as SL (associated more frequently with the verbs \emph{执行}/\emph{implement}, \emph{落实}/\emph{put into effect}, \emph{作出}/\emph{make}), and 23 times if English is used as SL (associated with the verbs as \emph{make}, \emph{engage}, \emph{carry}) and \textsc{include-as} occurs 37 times in Chinese (associated with the verbs \emph{包括}/\emph{include}, \emph{纳入}/\emph{incorporate}，\emph{记载}/\emph{record}) and 23 times if English is used as SL（associated with the verbs \emph{cover}, \emph{contain}, \emph{include}).

\subsubsection{Frame divergence}
Verbs with different meanings should exhibit two defining characteristics: they should occupy the same structural position and possess different frames.
The idea of \emph{occupying the same structural position} is defined as the case in which the two verbs have the same function governing a similar argument structure and conveying a related meaning.
As an example, we can consider the following sentences:
\ex.
\a. The Republic of Korea\textsubscript{agent} 
\textcolor{red!50!.}{launched} 
teaching and correspondence courses\textsubscript{theme} in remote areas.
\b. 大韩民国\textsubscript{agent}在边远地区\textcolor{red!50!.}{开办}了
教学和函授课程\textsubscript{theme}。

\exg.[] 大韩民国 在 边远 地区 开办 了 教学 和 函授 课程。\\
    Korea in remote area begin (-ed) teaching and correspondence courses.\\

\noindent The two sentences share the same arguments (\emph{agent} and \emph{theme}) surrounding a verb, but their annotation differs, in English, we have \emph{launch} annotated with \textsc{ESTABLISH} whereas in Chinese we have \emph{开办}/\emph{begin} annotated with \textsc{BEGIN}.


This category is characterized by two main aspects: 
\begin{enumerate}[label=(\roman*)]
    \item the aligned verb in the SL can be translated in the TL using verbs with a different meaning;
    \item in SL a sentence can have multiple verbs used to describe distinct actions, while in TL only some of them are maintained with the same meaning and other are changed by incorporating an adjective or an additional verb.
\end{enumerate}

\noindent As an example, we can refer to the following parallel sentences:
\ex.
\a. We urge the Palestinian Government and the Palestinian President to take immediate steps to confront individuals and groups \textcolor{red!50!.}{conducting} and \textcolor{red!50!.}{planning} terrorist attacks.
\b. 我们敦促巴勒斯坦政府和巴勒斯坦总统立即采取步骤，正视正在\textcolor{red!50!.}{进行}和计划\textcolor{red!50!.}{进行}恐怖主义攻击的个人和团体。

\exg.[] 我们 敦促 巴勒斯坦 政府 和 巴勒斯坦 总统 立即 采取 步骤， 正视 正在 进行 和 计划 进行 恐怖主义 攻击 的 个人 和 团体。\\
    We urge Palestinian Government and Palestinian President immediate take steps, confront be conduct and plan conduct terrorism attack of individual and group.\\

\noindent The English sentence in the last example includes the progressive form of the verb \emph{conduct}, defined as \emph{CIRCULATE\_SPREAD\_DISTRIBUTE}.
However, in Chinese, an isolated language with a simpler morphology compared to English, it is only possible to convey continuous tense by adding adverbs, e.g., \emph{正在} (\emph{in process of}) to express an ongoing action.
Consequently, \emph{conducting} (defined as \emph{CIRCULATE\_SPREAD\_DISTRIBUTE}) is translated using two words: \emph{正在} (literally \emph{in process of}) + \emph{进行} (literally \emph{conduct}, defined as \emph{HAPPEN\_OCCUR}).
Chinese maintains a similar syntactic structure within the sentence to ensure linguistic fluency; therefore, \emph{planning}(defined as \emph{PLAN\_SCHEDULE}) is translated as \emph{计划} (literally \emph{plan}, defined as \emph{REQUIRE\_NEED\_WANT\_HOPE}) + \emph{进行} (literally \emph{go on, carry on}, defined as \emph{HAPPEN\_OCCUR}), thus creating the phenomenon of having two Chinese verbs corresponding to one English verb.

\begin{table*}
\centering
\caption{Most common verbs in category 2.}\label{tab:verbcat2}
\resizebox{.99\textwidth}{!}{
\begin{tabular}{lr|lr}

\textbf{EN verb} & \textbf{\#} & \textbf{ZH verb}  & \textbf{\#}\\\hline
Note (PERCEIVE) & {13} & 指出/indicate (SIGNAL\_INDICATE) & {8} \\
  &  & 注意/note (SEE) & {5} \\
Take (TAKE) & {12} & 采取/take (CARRY-OUT-ACTION) & {10}\\
  &  & 作出/make (CARRY-OUT-ACTION) & 2 \\

Adopt (FOLLOW\_SUPPORT\_SPONSOR\_FUND) & {7} & 通过/pass (APPROVE\_PRAISE) & {6} \\
  &  & {采取/take (CARRY-OUT-ACTION)} & {1}\\
{Develop (CREATE\_MATERIALIZE)} & {6} & {制定/formulate (ORDER)} & {3} \\
 & & {提出/propose (PROPOSE)} & {1} \\
 & & {发展/develop (TRY)} & {1} \\
 & & {建立/build (ESTABLISH)} & {1} \\
Enhance (AMELIORATE) & 3 & 增进/improve (INCREASE\_ENLARGE\_MULTIPLY) & 1 \\
 & & 增强/strengthen (STRENGTHEN\_MAKE-RESISTANT) & 1 \\
 & & 加强/strengthen (STRENGTHEN\_MAKE-RESISTANT) & 1 \\
 Call (SUMMON) & 3 & 吁请/appeal (ASK\_REQUEST) & 2 \\
 & & 呼吁/appeal (ASK\_REQUEST) & 1 \\
Enhance (INCREASE\_ENLARGE\_MULTIPLY) & 2 & 扩大/expand (AMELIORATE) & 1 \\
 & & 加强/strengthen (STRENGTHEN\_MAKE-RESISTANT) & 1 \\
Call (ASK\_REQUEST) & 1 & 要求/request (REQUIRE\_NEED\_WANT\_HOPE) & 1 \\
\end{tabular}
}
\end{table*}

We also compared the frequency of verb occurrences in this category for both languages, and the results are presented in Table \ref{tab:verbcat2}.
We observed 
that a single verb in one language can have multiple translations in another language.


From Table \ref{tab:verbcat2}, it can be observed that the verb \emph{note} annotated with the frame \textsc{perceive} has two verb correspondences in Chinese.
An analysis of the parallel sentences indicates that this verb is translated as \emph{指出}/\emph{indicate} {8} times and as \emph{注意}/\emph{see} {5} times.
Such differences may stem from subtle distinctions in verb usage.
Below are two sentences illustrating the replacement of \emph{note} with the verbs \emph{指出}/\emph{indicate} and \emph{注意}/\emph{see}, respectively:

\ex.
\a. Many Governments \textcolor{red!50!.}{noted} achievements in women's health, including women's reproductive health.
\b. 许多国家的政府\textcolor{red!50!.}{指出}在妇女保健，包括妇女的生殖健康方面所取得的进展。\hfill

\exg.[] 许多 国家 的 政府 指出 在 妇女 保健， 包括 妇女 的 生殖 健康 方面 所取得 的 进展。\\ 
    Many country of government indicate in women health, include women of reproductive health field achieve of progress.\\

\ex.
\a.  The Working Group further \textcolor{red!50!.}{notes} that the Government has informed the Group that Ansar Mahmood and Sadek Awaed were ...
\b. 工作组进一步\textcolor{red!50!.}{注意}到，该国政府已经通知工作组 Ansar Mahmood 和 Sadek Awaed ...

\exg.[] 工作 组 进一步 注意 到， 该 国 政府 已经 通知 工作 组 Ansar Mahmood 和 Sadek Awaed ...\\ 
    Work group further notice to this country government already inform work group Ansar Mahmood and Sadek Awaed ...  \\

\noindent In the two English sentences, the frame and sense of \emph{note} are identical, specifically \textsc{perceive} and bn:00090661v, defined as \emph{Notice or perceive}.
However, in the two Chinese sentences, the frame and sense of \emph{指出}/\emph{indicate} are \textsc{signal\_indicate} and bn:00086679v defined as \emph{Indicate a place, direction, person, or thing; either spatially or figuratively}; whereas the frame and sense of \emph{注意}/\emph{see} are \textsc{see} and bn:00091096v defined as \emph{Observe with care or pay close attention to}.
From the two examples above, we can discern that the translation of \emph{note} into Chinese has different connotations.
\emph{注意}/\emph{see} tends to emphasize the active action of an individual, signifying that people intentionally focus their attention on a specific object or issue, as seen in the second example where the working group's focus is on \emph{release} and \emph{deport}, which are specific and defined actions.
On the other hand, \emph{指出}/\emph{indicate} emphasizes conveying information or viewpoints to others, highlighting a detail within a broader concept, as illustrated in the first example concerning \emph{achievements} in the realm of \emph{reproductive health}.

\begin{table}[h]
\centering
\caption{Number of times that verbs (V) are replaced with nominal (NE) and non-nominal expressions (non-NE)}\label{tab:nonverbal}
\begin{tabular}{l|r|r}
\toprule
&\multicolumn{1}{l|}{\textbf{ZH as SL}} & \multicolumn{1}{l}{\textbf{EN as SL}} \\
      V $\rightarrow$ NE & {229} & {45}\\
      V $\rightarrow$ non-NE & {55} & {31}\\
      \bottomrule
\end{tabular}
\end{table}

\subsubsection{Non-verbal alignment}
During the statistical examination of the third category, non-verb forms predominantly manifested as noun forms.
As a consequence of this, we further sub-divided this category into two parts:
\begin{itemize}
    \item Verbs in SL are translated into nouns in TL.
    \item Verbs in SL are translated into forms different from verbs and nouns.
\end{itemize}
To this end, in Table \ref{tab:nonverbal}, we provide the number of times this phenomenon occurs in our corpus.
The reason for the phenomena highlighted by our second category of divergences are due to translation choices \citep{wein-schneider-2021-classifying} mainly related to verb significance and communication efficiency.
Non-verbal alignment between Chinese and English is the only category that can be identified also using WALS, particularly through the analysis on action nominal constructions (ANCs) \citep{wals}.
ANCs are constructions featuring a nominalized verb as the head, combined with participants in the action.
WALS categorizes Chinese as having "no action nominals” in ANCs, meaning verbs are typically used directly in sentences.
In contrast, English is valued as "mixed," indicating it uses both nominalized forms and other constructions.

\paragraph{Verbs into nominal expressions}
As we have seen in Table \ref{tab:nonverbal}, the use of English nouns to replace Chinese verbs is very frequent.
This phenomenon can be attributed to the importance that verbs play in the Chinese language.
As an example, we can refer to the complexity of verb types in this language.
\citep{li1990xiandai}, in fact, identified 17 different types of verbs in Chinese and further subdivided them into other 13 classes according to their relationship with objects.
Consequently, when translating English nouns into Chinese, they are often converted into verb forms to adhere to the natural tendency of Chinese to emphasize verbs.
For example:
\ex.
\a. I have the honour to forward herewith the text of a declaration entitled: "Declaration on \textcolor{green!50!.}{Promoting} the universalization of the Convention on the \textcolor{red!50!.}{Prohibition} of the \textcolor{red!50!.}{Use, Stockpiling, Production} and \textcolor{red!50!.}{Transfer} of Anti-Personnel Mines and on Their \textcolor{red!50!.}{Destruction}" that was adopted ...
\b. 我谨随信转交...通过的一项宣言的全文，题为"关于\textcolor{green!50!.}{促进}各国普遍\textcolor{red!50!.}{加入}《关于\textcolor{red!50!.}{禁止使用、储存、生产}和\textcolor{red!50!.}{转让}杀伤人员地雷及\textcolor{red!50!.}{销毁}此种地雷的公约》的宣言"。

\exg.[] 我 谨 随 信 转交 ... 通过 的 一项 宣言 的 全文， 题 为 "关于 促进 各 国 普遍 加入 《关于 禁止 使用、 储存、 生产 和 转让 杀伤 人员 地雷 及 销毁 此 种 地雷 的 公约》 的 宣言"。\\ 
   I sincerely with letter transfer ... pass of one declaration of fullpage title is "about promote every country universalise join "about prohibit use, store, produce and transfer kill person mine and destory this type mine of convention" of declaration.    \\

\noindent As it is possible to see from these sentences, many verbs in Chinese can be aligned with nouns in English.
Another explanation for the phenomenon of non-verbal alignment can be associated with communicative efficiency.
Verbalizing English nouns in Chinese can enhance communicative efficiency.
Using verb forms, translated Chinese sentences or phrases can become more concise and expressive.
This linguistic efficiency in Chinese translation contributes to the preference for conversions from nouns to verbs, for example:
\ex.
\a. The omnibus resolution \textcolor{green!50!.}{emphasised} the unequivocal \textcolor{red!50!.}{undertaking} to accomplish the total \textcolor{red!50!.}{elimination} of nuclear weapons and the \textcolor{red!50!.}{implementation} of the 13 practical steps to \textcolor{green!50!.}{achieve} nuclear disarmament.
\b. 这项总括性的决议\textcolor{green!50!.}{强调}各国明确\textcolor{red!50!.}{保证}要实现彻底\textcolor{red!50!.}{消除}核武器和\textcolor{red!50!.}{执行} 13 个实际步骤， 以\textcolor{green!50!.}{实现}核裁军。

\exg.[] 这项 总括性 的 决议 强调 各 国 明确 保证 要 实现 彻底 消除 核 武器 和 执行 13个 实际 步骤， 以 实现 核 裁军。\\  
This omnibus of resolution emphasise every country clear ensure to achieve entirely eliminate nuclear weapon and implement 13 practical steps, to achieve nuclear disarmament.\\

\noindent In the proposed sentences, the terms \emph{保证}/\emph{guarantee}, \emph{消除}/\emph{eliminate}, and \emph{执行}/\emph{implement} are expressed as nouns in English.
The verb + NP structure is commonly used in the Chinese language, as in the following example:

\ex.
\a. 保证 + 要实现彻底消除核武器和执行 13 个实际步骤

\exg.[] 保证 + 要 实现 彻底 消除 核 武器 和 执行 13 个 实际 步骤\\  
guarantee + require achieve entirely eliminate nuclear weapon and implement 13 CL practical step\\

\ex.
\a. 消除 + 核武器

\exg.[] 消除 + 核 武器\\  
eliminate + nuclear weapon\\

\ex.
\a. 执行 + 13 个实际步骤

\exg.[] 执行 + 13 个 实际 步骤\\  
implement + 13 CL practical step\\

\noindent In these passages, verbs have been transformed into noun forms, even if they could have been expressed as:

\ex.
\a. 要实现彻底消除核武器和执行 13 个实际步骤 + 的 + 承诺

\exg.[] 要 实现 彻底 消除 核 武器 和 执行 13 个 实际 步骤 + 的 + 承诺\\  
require achieve entirely eliminate nuclear weapon and implement 13 CL practical step + of + promise\\

\ex.
\a. 核武器 + 的 + 消除

\exg.[] 核 武器 + 的 + 消除\\
nuclear weapon + of + elimination\\

\ex.
\a. 13 个实际步骤 + 的 + 执行

\exg.[] 13 个 实际 步骤 + 的 + 执行\\  
13 CL practical step + of + implementation\\

\noindent However, it's worth noting that in Chinese, when using the same nominalized form as in English, additional conjunctions are often added.
This can make sentences longer and more redundant, potentially affecting comprehension.

\begin{figure}[!ht]
\begin{center}
\includegraphics[width=0.4\linewidth,trim={.5cm .5cm .5cm .5cm},clip]{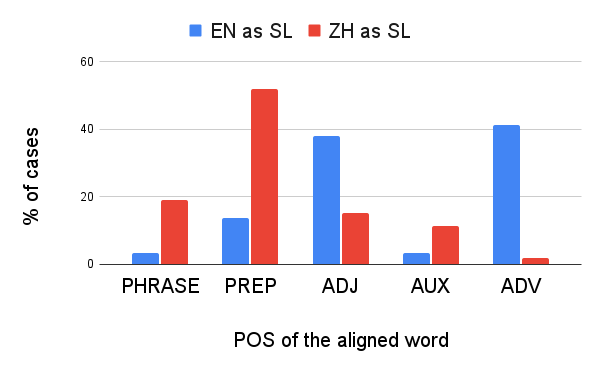} 
\caption{Verbs translated using different types of non-nominal expressions in both languages.}
\label{fig:nonverbal}
\end{center}
\end{figure}

\paragraph{Verbs into non-nominal expressions}
The statistics for non-nominal expressions are presented in Table \ref{tab:nonverbal}.
It shows that this phenomenon is less frequent than V $\rightarrow$ NE in both languages.
However, if Chinese is used as SL the decrease percentage of this phenomenon is around ${80\%}$ while if English is used as SL the decrease is just around ${59\%}$.
This indicates that there is a big disproportion between the two languages in the translation of verbal structures.
In Figure \ref{fig:nonverbal} we present the distribution of non-nominal divergences over five different cases, including phrases (i.e., the verb in the SL is translated using more than a word), prepositions, adjectives, auxiliaries, and adverbs.
We can see that Chinese verbs are mainly translated using prepositions, while English verbs are translated mainly using adjectives and adverbs.


\begin{table*}[ht]
\caption{Chinese verbs that have been translated using non-nominal and non-verbal expressions. The number of times this phenomenon happens for each verb is reported in parenthesis.)}
\resizebox{1\textwidth}{!}{
\begin{tabular}{llllll}
\toprule
\textbf{Type} & \textbf{Verbs}\\
      Phrase & 代表/\emph{represent} (4) & 没有/\emph{don't have} (3) & 考虑/\emph{consider} (1) & 脱贫/\emph{get rid of poverty} (1) & 阻碍/\emph{hinder} (1)\\
      Prep. & 包括/\emph{include} ({19}) & 为/\emph{become} (2) & 用于/\emph{use}（2）& 具有/\emph{have} (1) & 纳入/\emph{incorporate}（1）\\
      Adj. & 通过/\emph{pass} (1) & 调整/\emph{adjust}（1） & 意识/\emph{realize}（1）& 预计/\emph{estimate} (1) & 剥夺/\emph{deprive}（1）\\
      Aux. & 是/\emph{be} (3) & 属于/\emph{belong to} (2) & 感到/\emph{feel}（1）\\
      Adv. & 时断时续/\emph{discontinuous}（1)\\
      \bottomrule
\end{tabular}}
\end{table*}

\begin{table*}[ht]
\centering
\caption{English verbs that have been translated using non-nominal and non-verbal expressions. The number of times this phenomenon happens for each verb is reported in parenthesis.}
\resizebox{.7\textwidth}{!}{
\begin{tabular}{llllll}
\toprule
\textbf{Type} & \textbf{Verb}\\
Phrase & Marginalize (1)\\
Prep. & Relate (1) & Formalize (1) & Regard (1)& Base (1)\\
Adj. & Differ (2) & Highlight(1) & Commit (1)& Optimize (1) & Diversify（1）\\
Aux. & Enable（{3}）\\
Adv. & Continue（{3}）& Remain (2) & {Aim(2)} & Stay(1) &  Agree(1)\\
\bottomrule
\end{tabular}}
\end{table*}

\noindent Specific examples of these cases are presented below.

\ex.
\a.  Mr. Cheibani (Mali) (\textcolor{green!50!.}{spoke} in French): Despite the progress \textcolor{green!50!.}{made}, \textcolor{green!50!.}{curbing} the proliferation of, and illicit trafficking in, small arms and light weapons \textcolor{red!50!.}{continues} to be a major concern.
\b. 谢巴尼先生(马里)(\textcolor{red!50!.}{以}法语\textcolor{green!50!.}{发言}):尽管\textcolor{green!50!.}{取得}了进展，但\textcolor{green!50!.}{抑制}小武器和轻武器的扩 散和非法贩运仍然是一个主要关切事项。

\exg.[] 谢巴尼 先生 (马里) (以 法语 发言): 尽管 取得 了 进展， 但 抑制 小 武器 和 轻 武器 的 扩散 和 非法 贩运 仍然 是 一个 主要 关切 事项。\\  
XieBangni Mr. (Mali) (use French speak), despite make already progress, but inhibite small arms and light weapons of  proliferation and illegal trafficking still is a major concern matter.\\

\noindent In English, \emph{continue} (annotated with the frame \textsc{CONTINUE}) is a verb, while in Chinese, it is expressed using the adverb \emph{仍然}/\emph{still}.
It is important to note that \emph{continue} and \emph{仍然}/\emph{still} differ in their lexical meaning, as \emph{仍然}/\emph{still} is often translated as \emph{still}, but when combined with the rest of the sentence, it can be interpreted with a similar meaning.

The structure \emph{是}/\emph{be}+adj. or adv.+的/\emph{auxiliary word} is a typical expressive method in Chinese.
For example, \emph{the difference between A and B} can be expressed as \emph{A 和/and B 是/be 不同的/different}. 

Another example, considering Chinese as SL is reported below.
\ex.
\a. In other countries, such as in Lebanon, the leave period \textcolor{red!50!.}{differed} between the private sector (seven weeks) and the public sector (60 days).
\b. 在黎巴嫩等其他国家，私营部门(七周)和公共部门(60 天)的育儿假假期长度\textcolor{red!50!.}{不同}。

\exg.[] 在 黎巴嫩 等 其他 国家， 私营 部门 (七 周) 和 公共 部门 (60 天) 的 育儿 假 假期 长度 不同。\\  
   In Lebanon etc. other countries,  private sector (seven weeks) and  public sector (60 days) of child-raising vacation vacation duration different.\\

\noindent In this example, \emph{不同} /\emph{different} corresponds to the verb \emph{differ} in English.
This structure in Chinese can explain why many English verbs are translated into Chinese as adjectives or adverbs.
However, in sentence construction, due to considerations of sentence length, the 的/\emph{auxiliary word} in structure \emph{是}/\emph{be}+adj. or adv.+的/\emph{auxiliary word} is often omitted, as in the example above, where in Chinese, it is expressed more concisely as \emph{a} \emph{和}/\emph{and} \emph{b} (are) \emph{不同}/\emph{different}. 
Since \emph{是}/\emph{be} is a verb, the omission of the \emph{是}/\emph{be}+adj. or adv.+的/\emph{auxiliary word} structure can result in a discrepancy in the number of verbs in sentences between Chinese and English.

\begin{table}[!ht]
\caption{Number of times a defined verb is not translated in the parallel sentence}\label{tab:misverbs}
\begin{tabular}{llll}
\toprule
\textbf{English} & \textbf {\#}& \textbf{Chinese} & \textbf {\#}\\
Make & {5} & 有/\emph{have} & {9}\\
Wish & 4 & 进行/\emph{conduct} & {8}\\
Be & {4} & 是/\emph{be} & {5}\\
Have & 3 & 受到/\emph{receive} & 3\\
Exchange & 2& 获得/\emph{achieve} & 3\\
Bring & 1 & 开展/\emph{carry out} & 3\\
Deem & 1 & 得到/\emph{get} & 2\\
\bottomrule
\end{tabular}
\end{table}

\subsubsection{Misalignment}
Regarding the fourth class, we delved into potential commonalities among untranslated verbs.
We identified specific sentence structures that either reduced the use of verbs without compromising semantic expression or increased the use of verbs to align with grammar and language expression norms.

The number of Chinese verbs that do not have a corresponding translation in English is significantly higher than that of English verbs that do not have a translation in Chinese.
Specifically, this phenomenon happens {137} times when Chinese is the SL and {52} when English is the SL.
In Table \ref{tab:misverbs} we listed the most common verbs that do not have a corresponding translation in the other language, also including the number of times this phenomenon happens for each verb.

Similar to the reason for replacing verbs with non-verbal expressions, some of the verbs were left untranslated to avoid lengthy sentences caused by complex structures.
However, reasons for verbs not being translatable are more often associated with specific structural usage, primarily involving two structures: verb-noun and verb-NP-verb.
These two cases are presented below.

\paragraph{Verb-Noun Structure}

Regarding the verb-noun structure, the noun part often corresponds to the verb in the coordinate sentence.
The verb in the verb part is typically a light verb: 
a verb with little semantic content.
During the translation process, there exists a conversion from regular verbs to light verbs, as in the following example.

\ex.
\a.  ...the international community must \textcolor{red!50!.}{cooperate} at national, regional and international levels, and \textcolor{red!50!.}{exert} individual and collective efforts in relation to ...
\b. ...国际社会必须在国家、区域和国际各级\textcolor{red!50!.}{开展}合作，在...方面单独或者集体\textcolor{red!50!.}{作出}努力。

\exg.[] ...国际 社会 必须 在 国家、 区域 和 国际 各级 开展 合作， 在...方面 单独 或者 集体 作出 努力。\\ 
   ...International community must at national,  regional,  and international levels carry-out cooperation,  in...aspect individual or collective make efforts.\\
 
\noindent In this example, \emph{开展}/\emph{carry out} and \emph{作出}/\emph{make} are light verbs, and the real actions are expressed by the nouns (\emph{合作}/\emph{cooperation} and \emph{努力}/\emph{effort}).
Other common light verbs also presented in Table \ref{tab:misverbs}, are 受到/\emph{receive}, 获得/\emph{achieve} and 得到/\emph{get}.
It is not just a matter of language expression habits involving light verbs; the consistency in sentence structure between the preceding and following sentences can also lead to the use of light verbs, as in the following example.

\ex.
\a.  ...an interagency working group... with a special focus on the development and harmonization of methods, concepts and standards, coordination of data collection and training.  
\b. ...该工作组应特别\textcolor{green!50!.}{注重}\textcolor{red!50!.}{发展}和\textcolor{red!50!.}{统一}方法、概念和标准，\textcolor{red!50!.}{协调}数据收集工作并\textcolor{red!50!.}{进行}培训。 

\exg.[] ...该 工作 组 应 特别 注重 发展 和 统一 方法、 概念 和 标准， 协调 数据 收集 工作 并 进行 培训。\\ 
   ...this working group should especially emphasize develop and unify  method, concept, and standard, coordinate data collection work, and conduct training.\\

\noindent In the provided sentence, the use of the "verb + noun" structure in Chinese (发展/\emph{develop}和/\emph{and}统一/\emph{unify} + 方法/\emph{method}、概念/\emph{concept}和/\emph{and}标准/\emph{standard} and 协调/\emph{coordinate} + 数据/\emph{develop}收集/\emph{collection}工作/\emph{work})  contributes to the unity and coherence of the sentence.
Using a single verb form 培训/\emph{train} could compromise the equivalence and clarity of the subject and object in the sentence.
Therefore, the use of the expression 进行/\emph{conduct} + 培训/\emph{training} is more appropriate and precise in conveying the structure and meaning of the sentence.

\paragraph{Verb-NP-Verb Structure}
For the verb-NP-verb structure, typical in these parallel sentences for this quantity, the structure of the sentence containing the word 有/\emph{have} is prominent, occurring 15 times, and can be classified into two different ways:

\begin{itemize}
    \item 有/\emph{have} + NP + verb
    \item Verb + 有/\emph{have} + number + measure word (+ NP)
\end{itemize}

\noindent These cases are exemplified by the following parallel sentences.

\ex.
\a. Only 11 responding countries (Argentina, Austria, Belgium, Costa Rica, Cuba, Denmark, Finland, Germany, Iceland, the Netherlands and Norway) had \textcolor{red!50!.}{reached}...
\b. 仅仅\textcolor{red!50!.}{有} 11 个提出答复的国家(阿根廷、奥地利、比利时、哥斯达黎加、古巴、丹麦、芬 兰、德国、冰岛、荷兰和挪威)\textcolor{red!50!.}{达到}了...

\exg.[] 仅仅 有 11个 提出 答复 的 国家 (阿根廷、 奥地利、 比利时、 哥斯达黎加、 古巴、 丹麦、 芬兰、 德国、 冰岛、 荷兰 和 挪威) 达到了...\\ 
Only-just have 11 submit reply of countries (Argentina, Austria, Belgium, Costa-Rica, Cuba, Denmark, Finland, Germany, Iceland,  Netherlands, and Norway) reached...\\

\ex.
\a. Of the newly reported cases, 43 allegedly occurred in 2003, and \textcolor{red!50!.}{relate} to... 
\b. 在新报告的案件中，据称 2003 年发生的\textcolor{red!50!.}{有} 43 件，\textcolor{red!50!.}{涉及}...

\exg.[] 在 新 报告 的 案件 中， 据 称 2003 年 发生 的 有 43 件， 涉及...\\ 
In new report of cases in, according claims 2003 year happen of have 43 cases, involve...\\

\noindent In both mentioned examples, the verb 有/\emph{have} plays a specific role that cannot be aligned in parallel sentences and remains untranslatable.

Another characteristic structure is "是/\emph{be} + NP" in Chinese, which occurs six times.

\ex.
\a. The Committee \textcolor{red!50!.}{regrets} that the State party does not have...
\b. 委员会感到遗憾的\textcolor{red!50!.}{是}，缔约国没有...

\exg.[] 委员会 感到 遗憾 的 是， 缔约国 没有...\\ 
   Committee feels regretful of is, contracting-parties not..\\

\noindent It differs from the previously mentioned \emph{是}/\emph{be}+adj. or adv.+的/\emph{auxiliary word} structure because in the latter, the nominal part ends with 的/\emph{auxiliary word}, transforming an adjective or an adverb into a noun. In the 是/\emph{be} + NP structure, this transformation does not occur.
Consequently, unlike the omission of the verb in other constructions, this 是/\emph{be} + NP structure cannot be omitted.

\paragraph{Verb repetition in Chinese}
During the comparison, we also noticed a phenomenon of extensive verb repetition in a small number of Chinese sentences.
There are two sub-categories in the following examples:

\ex.
\a. In Burkina Faso, contraceptive use \textcolor{green!50!.}{increased} from 8.32 per cent in 1995 to 14.48 per cent in 2002; in Egypt from 24 per cent in 1980 to 56 per cent in 2000; in Uzbekistan from 13 per cent in 1980 to 62.3 per cent in 2003. 
\b. 在布基纳法索，避孕药具的使用比例从 1995 年的 8.32\%\textcolor{green!50!.}{增加}到了 2002 年的 14.48\%; 在埃及，从 1980 年的 24\%\textcolor{red!50!.}{增加}到了 2000 年的 56\%;在乌兹别克斯坦，从 1980 年的 13\%\textcolor{red!50!.}{增加}到 2003 年的 62.3\%。

\exg.[] 在 布基纳 法索， 避孕 药具 的 使用 比例 从 1995 年 的 8.32\% 增加 到了 2002 年 的 14.48\%;  在 埃及， 从 1980 年 的 24\% 增加 到了 2000 年 的 56\%; 在 乌兹别克斯坦， 从 1980 年 的 13\% 增加 到 2003 年 的 62.3\%。\\  
In Burkina Faso, contraceptive method of usage proportion from 1995 year of 8.32\% increased to 2002 year of 14.48\%; in Egypt, from 1980 year of 24\% increased to 2000 year of 56\%; in Uzbekistan, from 1980 year of 13\% increased to 2003 year of 62.3\%.\\

\ex.
\a. The economic performance of small island developing States over the past decade has been \textcolor{red!50!.}{mixed}.
\b. 小岛屿发展中国家的经济表现在过去十年期间\textcolor{red!50!.}{有}好\textcolor{red!50!.}{有}坏。

\exg.[] 小 岛屿 发展中 国家 的 经济 表现 在 过去 十 年 期间 有 好 有 坏。\\ 
Small island developing countries of economic performance in past ten year duration have good have bad.\\

\noindent In the first example, the use of the \emph{from...to...} structure in English replaces the repetition of the verb \emph{increase} found in the Chinese language, resulting in an alignment difference between the two languages (only the first 增加/\emph{increase} is aligned).

In the second example, these two sentences exhibit a more remarkable similarity in meaning rather than syntactic identity.
The structure 有/\emph{have}... 有/\emph{have}... is widely prevalent in Chinese, often used to describe the two sides of a coin.
Another concrete example is 有/\emph{have}棱/\emph{edge}有/\emph{have}角/\emph{corner}, which is frequently employed to depict something structured, well-defined, and precise.
While this expression does not pose semantic comprehension issues, a discrepancy exists between the number of verbs and the meanings of verbs in English and Chinese.

\begin{table*}[!t]
\centering
\caption{Results of the X-SRL annotation projection reported as correct (C), false positive (FS), false negative (FN) precision (P), recall (R) and f1 measure (F1) for both languages and for verbs (V) and semantic roles. Calculated using xsrl\_mbert\_aligner.}\label{tab:annproj}
\resizebox{.9\textwidth}{!}{
\begin{tabular}{l rrrrrr | rrrrrr}
\toprule
& \multicolumn{6}{c|}{Chinese as SL} & \multicolumn{6}{c}{English as SL}\\
& \textbf{C} & \textbf{FP} & \textbf{FN} & \textbf{P} & \textbf{R} & \textbf{F1} & \textbf{C} & \textbf{FP} & \textbf{FN} & \textbf{P} & \textbf{R} & \textbf{F1} \\
\textbf{predicates} & 442 & 56 & 228 & 88.76 & 65.97 & 75.68 & 326 & 37 & 565 & 89.81 & 36.59 & 51.99 \\
agent & 103 & 87 & 213 & 54.21 & 32.59 & 40.71 & 64 & 39 & 312 & 62.14 & 17.02 & 26.72 \\
asset & 0 & 1 & 1 & 0 & 0 & 0 & 0 & 0 & 1 & 0 & 0 & 0 \\
attribute & 0 & 12 & 22 & 0 & 0 & 0 & 0 & 4 & 34 & 0 & 0 & 0 \\
beneficiary & 5 & 15 & 28 & 25 & 15.15 & 18.87 & 4 & 6 & 51 & 40 & 7.27 & 12.31 \\
cause & 0 & 1 & 2 & 0 & 0 & 0 & 0 & 1 & 2 & 0 & 0 & 0 \\
co-agent & 0 & 2 & 10 & 0 & 0 & 0 & 0 & 0 & 6 & 0 & 0 & 0 \\
co-patient & 0 & 1 & 3 & 0 & 0 & 0 & 0 & 0 & 3 & 0 & 0 & 0 \\
co-theme & 3 & 6 & 12 & 33.33 & 20 & 25 & 1 & 1 & 20 & 50 & 4.76 & 8.7 \\
destination & 2 & 10 & 14 & 16.67 & 12.5 & 14.29 & 0 & 3 & 24 & 0 & 0 & 0 \\
experiencer & 1 & 5 & 25 & 16.67 & 3.85 & 6.25 & 0 & 6 & 18 & 0 & 0 & 0 \\
extent & 0 & 3 & 5 & 0 & 0 & 0 & 0 & 0 & 8 & 0 & 0 & 0 \\
goal & 4 & 35 & 47 & 10.26 & 7.84 & 8.89 & 2 & 15 & 92 & 11.76 & 2.13 & 3.6 \\
instrument & 0 & 7 & 7 & 0 & 0 & 0 & 0 & 1 & 16 & 0 & 0 & 0 \\
location & 0 & 6 & 12 & 0 & 0 & 0 & 0 & 2 & 14 & 0 & 0 & 0 \\
material & 0 & 2 & 0 & 0 & 0 & 0 & 0 & 0 & 3 & 0 & 0 & 0 \\
patient & 21 & 54 & 110 & 28 & 16.03 & 20.39 & 15 & 21 & 182 & 41.67 & 7.61 & 12.88  \\
product & 0 & 2 & 4 & 0 & 0 & 0 & 0 & 2 & 6 & 0 & 0 & 0 \\
recipient & 8 & 14 & 31 & 36.36 & 20.51 & 26.23 & 5 & 13 & 50 & 27.78 & 9.43 & 14.08 \\
result & 1 & 7 & 20 & 12.5 & 4.76 & 6.9 & 0 & 3 & 23 & 0 & 0 & 0 \\
source & 2 & 8 & 18 & 20 & 10 & 13.33 & 0 & 6 & 22 & 0 & 0 & 0 \\
stimulus & 1 & 4 & 25 & 20 & 3.85 & 6.45 & 0 & 7 & 21 & 0 & 0 & 0 \\
theme & 57 & 106 & 268 & 34.97 & 17.54 & 23.36 & 25 & 54 & 347 & 31.65 & 6.72 & 11.09 \\
time & - & - & - & - & - & - & 0 & 0 & 1 & 0 & 0 & 0 \\
topic & 10 & 30 & 53 & 25 & 15.87 & 19.42 & 9 & 8 & 85 & 52.94 & 9.57 & 16.22 \\
value & 0 & 3 & 1 & 0 & 0 & 0 & 0 & 0 & 11 & 0 & 0 & 0 \\
\midrule
\textbf{Overall} & 660 & 477 & 1159 & 58.05 & 36.28 & 44.65 & 451 & 229 & 1917 & 66.32 & 19.05 & 29.59 \\
\bottomrule
\end{tabular}}
\end{table*}

\begin{table*}[!ht]
\centering
\caption{Results of the X-SRL annotation projection reported as correct (C), false positive (FS), false negative (FN) precision (P), recall (R) and f1 measure (F1) for both languages and for verbs (V) and semantic roles. Calculated using  awesome-align.}\label{tab:annproj_aws}
\resizebox{.9\textwidth}{!}{
\begin{tabular}{l rrrrrr | rrrrrr}
\toprule
& \multicolumn{6}{c|}{Chinese as SL} & \multicolumn{6}{c}{English as SL}\\
& \textbf{C} & \textbf{FP} & \textbf{FN} & \textbf{P} & \textbf{R} & \textbf{F1} & \textbf{C} & \textbf{FP} & \textbf{FN} & \textbf{P} & \textbf{R} & \textbf{F1} \\
\textbf{predicates} & 319 & 47 & 351 & 87.16 & 47.61 & 61.58 & 274  & 27  & 617  & 91.03  & 30.75  & 45.97 \\
agent & 205 & 129 & 176 & 61.38 & 53.81 & 57.35 & 212  & 116  & 297  & 64.63  & 41.65  & 50.66 \\
asset & 0 & 0 & 1 & 0.0 & 0.0 & 0.0 & 0  & 1  & 1  & 0.0  & 0.0  & 0.0 \\
attribute & 1 & 4 & 21 & 20.0 & 4.55 & 7.41 & 1  & 5  & 36  & 16.67  & 2.7  & 4.65 \\
beneficiary & 10 & 21 & 23 & 32.26 & 30.3 & 31.25 & 9  & 8  & 47  & 52.94  & 16.07  & 24.66 \\
cause & 1 & 1 & 1 & 50.0 & 50.0 & 50.0 & 1  & 1  & 1  & 50.0  & 50.0  & 50.0 \\
co-agent & 0 & 2 & 10 & 0.0 & 0.0 & 0.0 & 0  & 4  & 7  & 0.0  & 0.0  & 0.0 \\
co-patient & 0 & 2 & 3 & 0.0 & 0.0 & 0.0 & 0  & 3  & 3  & 0.0  & 0.0  & 0.0 \\
co-theme & 4 & 3 & 12 & 57.14 & 25.0 & 34.78 & 4  & 2  & 17  & 66.67  & 19.05  & 29.63 \\
destination & 1 & 4 & 15 & 20.0 & 6.25 & 9.52 & 1  & 2  & 23  & 33.33  & 4.17  & 7.41 \\
experiencer & 3 & 7 & 24 & 30.0 & 11.11 & 16.22 & 4  & 7  & 14  & 36.36  & 22.22  & 27.58 \\
extent & 1 & 3 & 4 & 25.0 & 20.0 & 22.22 & 0  & 2  & 8  & 0.0  & 0.0  & 0.0 \\
goal & 11 & 38 & 40 & 22.45 & 21.57 & 22.0 & 9  & 18  & 92  & 33.33  & 8.91  & 14.06 \\
instrument & 0 & 5 & 7 & 0.0 & 0.0 & 0.0 & 0  & 1  & 16  & 0.0  & 0.0  & 0.0 \\
location & 0 & 3 & 12 & 0.0 & 0.0 & 0.0 & 0  & 3  & 14  & 0.0  & 0.0  & 0.0 \\
material & 0 & 1 & 0 & 0.0 & 1.0 & 0.0 & 0  & 0  & 3  & 0.0  & 0.0  & 0.0 \\
patient & 50 & 36 & 84 & 58.14 & 37.31 & 45.45 & 48  & 26  & 165  & 64.86  & 22.54  & 33.45 \\
product & 1 & 1 & 3 & 50.0 & 25.0 & 33.33 & 1  & 1  & 5  & 50.0  & 16.67  & 25.0 \\
recipent & 0 & 1 & 0 & 0.0 & 1.0 & 0.0 & 0  & 0  & 2  & 0.0  & 0.0  & 0.0 \\
recipient & 15 & 16 & 24 & 48.39 & 38.46 & 42.86 & 12  & 7  & 42  & 63.16  & 22.22  & 32.87 \\
result & 3 & 14 & 18 & 17.65 & 14.29 & 15.79 & 3  & 7  & 21  & 30.0  & 12.5  & 17.65 \\
source & 0 & 9 & 20 & 0.0 & 0.0 & 0.0 & 0  & 5  & 25  & 0.0  & 0.0  & 0.0 \\
stimulus & 3 & 5 & 23 & 37.5 & 11.54 & 17.65 & 3  & 10  & 18  & 23.08  & 14.29  & 17.65 \\
theme & 95 & 131 & 251 & 42.04 & 27.46 & 33.22 & 90  & 88  & 335  & 50.56  & 21.18  & 29.85 \\
time & - & - & - & - & - & - & 0  & 0  & 1  & 0.0  & 0.0  & 0.0 \\
topic & 20 & 38 & 44 & 34.48 & 31.25 & 32.79 & 21  & 15  & 76  & 58.33  & 21.65  & 31.58 \\
value & 0 & 0 & 1 & 0.0 & 0.0 & 0.0 & 0  & 1  & 12  & 0.0  & 0.0  & 0.0 \\
\midrule
Overall & 743 & 521 & 1168 & 58.78 & 38.88 & 46.8 & 693  & 360  & 1898  & 65.81  & 26.75  & 38.04 \\
\bottomrule
\end{tabular}}
\end{table*}

\section{Annotation Projection Results}
To evaluate the impact of predicate-argument structure divergences, we performed an annotation projection experiment using the same set of sentences analyzed in this study. For this purpose, we utilized two annotation projection methods: X-SRL \citep{daza-frank-2020-x} and awesome-align \citep{dou-neubig-2021-word}.
X-SRL was used in previous work to create the homonymous silver dataset for SRL \citep{daza-frank-2020-x}.
It uses mBERT embeddings \citep{devlin-etal-2019-bert} to compute the pair-wise cosine similarity between source and target tokens, guiding word alignments according to this measure.
These measures are stored in a similarity matrix $SM$ with $p$ columns and $q$ rows that correspond to the word-pieces of the source sentence and the word-pieces of the target sentence, respectively.
From each column in $SM$, the $k$ most similar pairs are selected allowing in this way one-to-many mappings.
These mappings are then filtered, keeping only annotated tokens and verbs.
After this step, the candidate with the highest score is aligned.
If the list is empty, no projection is performed.
We used the S2T configuration of X-SRL in our experiments as recommended by the authors for high precision and high recall results.
It performs source-to-target-alignments using the part-of-speech and the SRL verb annotation to filter non-relevant word alignments.
Awesome-align is a word aligner system that was adapted by us to project SRL annotations between aligned texts following the same procedure described above for X-SRL.
It leverages mBERT by fine-tuning it on parallel text with alignment objectives.
The dot product between source and target contextualized embeddings $h_x$ and $h_y$ is computed to get the similarity matrix $S$.
A normalization function is used to convert the similarity matrix into values on the probability simplex $S_{xy} = N (S)$, and treat $S_{xy}$ as the source-to-target alignment matrix.
Optimal transport is then used to find a mapping that moves probability mass from the distribution of source to the distribution of the target in $S_{xy}$, minimizing the distance between alignments pairs.
Finally, bidirectional alignments are extracted by storing the intersection values of the two matrices $S_{xy}$ and $S_{yx}$ into a new matrix $A$.
Different objectives are also used to fine-tune the LM, including:
\begin{enumerate}[label=(\roman*)]
    \item Masked Language Modeling on the task data;
    \item Translation Language Modeling concatenating sentence in the source and the target languages and performing MLM on them;
    \item Self-training Objective (SO) on the alignments matrix $A$ encouraging words aligned in $A$ to have further closer contextualized representations.
    \item Parallel Sentence Identification;
    \item Consistency Optimization to explicitly encourage the consistency between the two alignment matrices $S_{xy}$ and $S_{yx}$.
\end{enumerate}

\subsection{Quantitative Analysis}
The results of this analysis are presented in Table \ref{tab:annproj} (X-SRL) and Table \ref{tab:annproj_aws} (awesome-align). These tables display the number of correct (C), false positive (FP), and false negative (FN) cases, along with precision (P), recall (R), and F1 score (F1) for both languages, categorized by verbs (V) and semantic roles.
As we can see, the overall scores are very low for both languages and for both techniques.
However, when Chinese is used as SL, the F1 score is much higher ($44.65$ with X-SRL and $44.8$ with awesome-align) than when English is used as SL ($29.59$ with X-SRL and $38.04$ with awesome-align).
This discrepancy is due to the fact that in Chinese there are many more annotations and for this reason when English is used as SL the recall of the model  is very low ($19.05$ with X-SRL and $26.04$ with awesome-align).
As a consequence, the fraction of FP cases is higher when Chinese is used as source language ($21\%$ with X-SRl and $27\%$ with awesome-align vs $9\%$ with X-SRL and $13\%$ with awsome-align).
On the other hand, the fraction of FN cases is higher when English is used as SL ($80\%$ with X-SRL and $73\%$ with awesome-align vs $64\%$ with X-SRL and $61\%$ awesome-align).
Although using English as the source language (SL) results in a lower F1 score, it is noteworthy that the model's precision is higher in this scenario ($66.32$ with X-SRL and $65.81$ with awesome-align) compared to when annotations are projected from Chinese to English ($58.05$ with X-SRL and $58.78$ with awesome-align). The high rate of false negatives (FN) observed in both languages, regardless of the projection model used, highlights the challenges in identifying suitable candidates for alignment.

Regarding \emph{predicates}, the analysis reveals a similar pattern. Notably, the F1 score is significantly higher when Chinese is used as the source language (SL): $75.68$ with X-SRL and $61.58$ with awesome-align, compared to $51.99$ with X-SRL and $45.97$ with awesome-align when English serves as the SL. These results are strongly influenced by the low recall observed in both annotation projection models.
Consequently, the results on all arguments are low as well.
This outcome occurs because, even if a model successfully aligns arguments, they cannot be projected unless they are properly attached to a predicate; otherwise, the result would be incomplete predicate-argument structures. Even for the most common semantic roles—\emph{agent}, \emph{patient}, and \emph{theme}—the F1 score with X-SRL does not exceed $40.71$ when Chinese is the source language (SL) and $26.72$ when English is the SL, compared to $57.35$ and $50.66$, respectively, using awesome-align. The high rate of false negatives (FN) in both languages further degrades the models' performance for semantic roles. Additionally, low-frequency semantic roles are rarely annotated correctly with X-SRL, especially when English is the SL. Overall, X-SRL outperforms awesome-align in predicate projection, while awesome-align achieves significantly better results in aligning semantic roles.

\begin{table*}[ht]
    \centering
    \caption{Frame divergence between ZN-gold and ZN-projected. Numbered by example (\#) and divided using divergences categories (div. cat.).}
    \resizebox{.8\textwidth}{!}{
    \begin{tabular}{rrlll}
    \toprule
        \textbf{\#} & \textbf{div. cat.} & \textbf{Verb} & \textbf{ZN-projected} & \textbf{ZN-gold}  \\
        1 & 2 & 希望/hope & CHOOSE & REQUIRE\_NEED\_WANT\_HOPE \\
        2 & 4 & 随信/enclosed herewith & EXIST\_WITH\_FEATURE & \\
        2 & 3 & 使用/use & & USE \\
        4 & 3 & 在/at & ESTABLISH & \\
        5 & 2 & 正视/face & OBLIGE\_FORCE & FACE\_CHALLENGE \\
        5 & 2 & (正在)进行/conduct & PLAN\_SCHEDULE & HAPPEN\_OCCUR \\
        6 & 2 & 指出/indicate & PERCEIVE & SIGNAL\_INDICATE \\
        7 & 3 & 到/to & PERCEIVE & \\
        7 & 3 & 被/passive marker& LIBERATE\_ALLOW\_AFFORD & \\
        7 & 3 & 解除/terminate& & EXEMPT \\
        8 & 3 & 随信/with letter & EXIST\_WITH\_FEATURE & \\
        8 & 3 & 使用/use & & USE \\
        19 & 2 & 认为/think& AGREE\_ACCEPT & SUBJECTIVE-JUDGING \\
        20 & 2 & 达到/reach& ACHIEVE & REACH \\
        21 & 2 & 是/be & REGRET\_SORRY & MATCH \\
        21 & 2 & 没有/do not have& EXIST-WITH-FEATURE & MISS\_OMIT\_LACK \\
    \bottomrule    
    \end{tabular}}
    \label{tab:ZN-gold_and_ZN-projected}
\end{table*}

\subsection{Qualitative Analysis}
We compared the frame annotations of the examples presented in the paper with the automatic annotations obtained using the annotation projection model presented above.
The main differences that we found and that reflect our divergences classes can be summarized as follows:
\begin{itemize}
\item Prepositions are often translated into verbs (category 3).
For example, \emph{在/at} is interpreted as a preposition in the structure of "在 + location/time + verb" while is interpreted as a verb meaning exist in the structure of "在 + location/noun (without a following verb)".
However, in sentence 4, \emph{在/at} in the structure of "在 + location/time + verb"  is annotated as a verb in ZH-projected.
In EN-projected we have the same problem.
For example, in sentence 6, \emph{including} is a preposition, but it is annotated as a verb in EN-projected.
\item Passive markers in passive sentences are annotated as verbs in projected with the frame of the actual verb in the sentence, while the actual verb is not annotated with frame.
In Chinese, the passive marker \emph{被 (being)} works as a particle in the structure of "N + 被 + V + NP" and as an adjective in the structure of "N1 + 被 + N2 + V".
In sentence 7, \emph{被 (being)} in \emph{被解除 (being terminate)}  is annotated as a verb by the projection. 
Also in EN-projected, in example 24, \emph{(have) been (mixed)} as a passive marker is annotated as a verb.
\item Negation in negative sentences is not reflected in Chinese.
For example, in sentence 21, \emph{没有 (do not have)} in ZN-gold is framed as EXIST-WITH-FEATURE, failing to indicate the negation of the verb, indicated by MISS\_OMIT\_LACK in ZH-gold.
\end{itemize}

In Table \ref{tab:ZN-gold_and_ZN-projected} and Table \ref{tab:EN-gold_and_EN-projected}, we can see some common problems.
The analysis of Table \ref{tab:ZN-gold_and_ZN-projected} reveals several problematic issues regarding the alignment between the ZN-projected and ZN-gold frame annotations. One major problem is inconsistency in frame selection, where the projected frames often diverge significantly from the gold standard, suggesting a lack of semantic alignment (e.g., "是/be" mapped to "REGRET\_SORRY" instead of the more neutral "MATCH"). Another issue is missing annotations on both sides: some entries have no corresponding frame in either ZN-projected or ZN-gold, indicating incomplete frame projection or gold standard labeling (e.g., rows with verbs like "使用/use" or "解除/terminate"). Additionally, the repetition of examples with different divergence categories for the same sentence number (e.g., \#2 and \#5) points to multi-layered divergence, where more than one type of mismatch occurs within the same sentence, potentially compounding interpretation errors. Finally, the presence of ambiguous or overly broad frames in the ZN-projected column (e.g., "LIBERATE\_ALLOW\_AFFORD" for a passive marker) suggests that the projection system may overgeneralize or misinterpret syntactic cues. Collectively, these issues highlight the challenges of accurate frame alignment in cross-lingual or automated projection contexts, underscoring the need for improved disambiguation and context-aware modeling.
The analysis of Table \ref{tab:EN-gold_and_EN-projected} comparing EN-projected and EN-gold frame annotations highlights several systematic issues in frame alignment and projection accuracy. First, semantic misalignment is prevalent, with projected frames often failing to capture the intended meaning in the gold annotations—for example, “prefer” is projected as “REQUIRE\_NEED\_WANT\_HOPE” instead of the more precise “CHOOSE,” and “regrets” is mismatched with “MATCH” instead of “REGRET\_SORRY.” These mismatches suggest a tendency of the projection system to opt for semantically adjacent but imprecise frames. Second, the table exhibits numerous instances of missing gold annotations (e.g., “including” and “been”), which can stem from either annotation gaps or projection limitations, further complicating evaluation. Third, several entries show inconsistent mappings for repeated verbs across examples, such as “promoting” being assigned the same incorrect frame “SPEED\_UP” in different contexts, indicating a lack of contextual adaptation. There’s also evidence of lexical misinterpretation, as seen in “adopted” being projected as “SEND” when “LEARN” was intended, revealing possible confusion between syntactic structure and event semantics. Altogether, these problems reveal weaknesses in projection quality and semantic sensitivity, emphasizing the need for more context-aware, lexically nuanced projection models in cross-lingual or automatic frame alignment tasks.

\begin{table*}[h]
    \centering
    \caption{Frame divergence between EN-gold and EN-projected. Numbered by example and divided using divergences categories (div. cat.)}
    \resizebox{.8\textwidth}{!}{
    \begin{tabular}{rrlll}
    \toprule
        \textbf{\#} & \textbf{div. cat.} & \textbf{Verb} & \textbf{EN-projected} & \textbf{EN-gold}. \\
        1 & 2 & prefer & REQUIRE\_NEED\_WANT\_HOPE & CHOOSE \\
        2 & 3 & promoting &  SPEED\_UP & \\
        2 & 2 & adopted &  SEND & LEARN \\ 
        5 & 2 & urge &  FACE\_CHALLENGE & OBLIGE\_FORCE \\
        5 & 2 & planning &  HAPPEN\_OCCUR & PLAN\_SCHEDULE \\
        6 & 2 & noted &  SIGNAL\_INDICATE & PERCEIVE \\
        6 & 3 & including &  INCLUDE-AS & \\
        8 & 3 & promoting &  SPEED\_UP & \\
        8 & 2 & adopted &  SEND & LEARN \\
        16 & 2 & spoke &  USE & SPEAK \\
        16 & 2 & continues &  REDUCE\_DIMINISH & CONTINUE \\
        18 & 2 & exert &  FACE\_CHALLENGE & CARRY-OUT-ACTION \\
        19 & 2 & agreed &  SUBJECTIVE\-JUDGING & AGREE\_ACCEPT \\
        19 & 2 & set &  ESTABLISH & PREPARE \\
        20 & 2 & reached &  REACH & ACHIEVE \\
        20 & 2 & have &  CREATE\_MATERIALIZE & INCITE\_INDUCE \\
        22 & 2 & regrets &  MATCH & REGRET\_SORRY \\
        22 & 2 & have & MISS\_OMIT\_LACK & EXIST-WITH-FEATURE \\
        24 & 3 & been & EXIST-WITH-FEATURE & \\
    \bottomrule        
    \end{tabular}}
    \label{tab:EN-gold_and_EN-projected}
\end{table*}

\begin{table}[!h]
\caption{Accuracy on predicate sense disambiguation on the test of the UniteD-SRL dataset in English (EN) and Chinese (ZH). The reported results are obtained using XLM-RoBERTa, finding a consistent. As reported in \citep{tripodi2021united}.}\label{tab:united_exp}
\centering
\resizebox{.3\linewidth}{!}{
\begin{tabular}{l r r }
\toprule
Lang train & \multicolumn{2}{c}{Lang test} \\ 
 & EN & ZH \\
\midrule
EN & \textbf{87.0} & 54.7 \\
ZH & 63.6 & \textbf{78.0} \\
\bottomrule
\end{tabular}}
\end{table}

\section{Cross-Lingual SRL Results}
\citep{tripodi2021united} conducted a cross-lingual SRL experiment in which a state-of-the-art SRL model \cite[CN20]{conia-navigli-2020-bridging} was used to train  two monolingual SRL annotators for English and Chinese.
These models were then tested in a zero-shot fashion on the language they were not trained.
From the results of this experiment, reported for completeness in Table \ref{tab:united_exp}, it emerged clearly that the models are not able to perform cross-lingual transfer and that the two models trained on the same parallel sentences have extremely different performances (the model trained on English achieve 9 F1 points more that the model trained on Chinese).
The study conducted in this paper allows to explain why these phenomena occur.
In particular, in light of our analysis, it is possible to identify the reason behind the lower performances of the Chinese model that are mainly related to the higher complexity, in term of predicate-argument structures, of the Chinese language.
It is also possible to identify the reasons behind the low zero-shot performances, related mainly on predicate-argument structure divergences.
The drop in performances of the Chinese model is, in fact, lower that the drop reported by the English model ($-18.5\%$ and $-37.1\%$, respectively).
This result is in line with the results reported in our annotation projection experiments (see Table \ref{tab:annproj} and Table \ref{tab:annproj_aws}).

\section{Conclusion}
In this study, we conducted an extensive analysis of predicate-argument structures in a corpus of 400 professionally translated parallel sentences in English and Chinese, uncovering significant differences between the two languages. One of the most notable distinctions is the overall number of verbs: Chinese exhibits 25\% more verbal structures, which in some cases correspond to non-verbal (mainly nominal) structures in English, while in other cases, the verbs are entirely absent.
Beyond this general observation, our analysis identified four alignment categories for predicates: frame convergence, frame divergence, non-verbal alignment, and misalignment. These categories show distinct distributions depending on which language is used as the source language (SL). With the higher number of predicates in Chinese, alignments (58\% of cases, category 1) and frame divergences (category 2) are more easily identified when English serves as the SL. Conversely, when Chinese is the SL, the alignment rate drops to 43.4\%, and the other divergence categories—particularly non-verbal alignments (category 3)—occur more frequently. This highlights that the two languages often employ different syntactic structures to convey similar meanings.
Additionally, our analysis reveals that for the same English verb, various Chinese verbs may be used, and untranslated verbs are significantly more common when Chinese serves as the SL. To complement our findings with empirical evidence, we used the parallel corpus as a ground-truth dataset for an annotation projection experiment. This experiment demonstrated a notable asymmetry in outcomes depending on the SL. When English is the SL, the F1 score for correctly aligned predicates is low (51.99), whereas it is significantly higher (75.68) when Chinese is the SL.
This analysis also informed the results of a cross-lingual SRL experiment in a zero-shot setting, showing how linguistic divergences obstruct effective language transfer. Such insights were only possible through the systematic examination and categorization of predicate-argument structure divergences.
These findings underscore the necessity of carefully selecting the source language for annotation or model adaptation in cross-lingual transfer learning. While this selection is empirically challenging due to the predominance of Indo-European language resources, we argue that it is a critical prerequisite for validating any scientific claims in cross-lingual NLP.

\bibliographystyle{plainnat}
\bibliography{bib}

@inproceedings{DBLP:conf/lrec/ZiemskiJP16,
  author       = {Michal Ziemski and
                  Marcin Junczys{-}Dowmunt and
                  Bruno Pouliquen},
  editor       = {Nicoletta Calzolari and
                  Khalid Choukri and
                  Thierry Declerck and
                  Sara Goggi and
                  Marko Grobelnik and
                  Bente Maegaard and
                  Joseph Mariani and
                  H{\'{e}}l{\`{e}}ne Mazo and
                  Asunci{\'{o}}n Moreno and
                  Jan Odijk and
                  Stelios Piperidis},
  title        = {The United Nations Parallel Corpus v1.0},
  booktitle    = {Proceedings of the Tenth International Conference on Language Resources
                  and Evaluation {LREC} 2016, Portoro{\v{z}}, Slovenia, May 23-28, 2016},
  publisher    = {European Language Resources Association {(ELRA)}},
  year         = {2016},
  url          = {http://www.lrec-conf.org/proceedings/lrec2016/summaries/1195.html}
}

@inproceedings{yao-koller-2022-structural,
    title = "Structural generalization is hard for sequence-to-sequence models",
    author = "Yao, Yuekun  and
      Koller, Alexander",
    editor = "Goldberg, Yoav  and
      Kozareva, Zornitsa  and
      Zhang, Yue",
    booktitle = "Proceedings of the 2022 Conference on Empirical Methods in Natural Language Processing",
    month = dec,
    year = "2022",
    address = "Abu Dhabi, United Arab Emirates",
    publisher = "Association for Computational Linguistics",
    url = "https://aclanthology.org/2022.emnlp-main.337/",
    doi = "10.18653/v1/2022.emnlp-main.337",
    pages = "5048--5062"
}

@book{wals,
  address   = {Leipzig},
  editor    = {Matthew S. Dryer and Martin Haspelmath},
  publisher = {Max Planck Institute for Evolutionary Anthropology},
  title     = {WALS Online},
  url       = {https://wals.info/},
  year      = {2013}
}

@inproceedings{blevins-zettlemoyer-2022-language,
    title = "Language Contamination Helps Explains the Cross-lingual Capabilities of {E}nglish Pretrained Models",
    author = "Blevins, Terra  and
      Zettlemoyer, Luke",
    editor = "Goldberg, Yoav  and
      Kozareva, Zornitsa  and
      Zhang, Yue",
    booktitle = "Proceedings of the 2022 Conference on Empirical Methods in Natural Language Processing",
    month = dec,
    year = "2022",
    address = "Abu Dhabi, United Arab Emirates",
    publisher = "Association for Computational Linguistics",
    pages = "3563--3574",
}

@inproceedings{conia-navigli-2020-bridging,
    title = "Bridging the Gap in Multilingual Semantic Role Labeling: a Language-Agnostic Approach",
    author = "Conia, Simone  and
      Navigli, Roberto",
    editor = "Scott, Donia  and
      Bel, Nuria  and
      Zong, Chengqing",
    booktitle = "Proceedings of the 28th International Conference on Computational Linguistics",
    month = dec,
    year = "2020",
    address = "Barcelona, Spain (Online)",
    publisher = "International Committee on Computational Linguistics",
    pages = "1396--1410",
}

@inproceedings{dou-neubig-2021-word,
    title = "Word Alignment by Fine-tuning Embeddings on Parallel Corpora",
    author = "Dou, Zi-Yi  and
      Neubig, Graham",
    editor = "Merlo, Paola  and
      Tiedemann, Jorg  and
      Tsarfaty, Reut",
    booktitle = "Proceedings of the 16th Conference of the European Chapter of the Association for Computational Linguistics: Main Volume",
    month = apr,
    year = "2021",
    address = "Online",
    publisher = "Association for Computational Linguistics",
    pages = "2112--2128",
}

@article{ponti-etal-2019-modeling,
    title = "Modeling Language Variation and Universals: A Survey on Typological Linguistics for Natural Language Processing",
    author = "Ponti, Edoardo Maria  and
      O{'}Horan, Helen  and
      Berzak, Yevgeni  and
      Vuli{\'c}, Ivan  and
      Reichart, Roi  and
      Poibeau, Thierry  and
      Shutova, Ekaterina  and
      Korhonen, Anna",
    journal = "Computational Linguistics",
    volume = "45",
    number = "3",
    month = sep,
    year = "2019",
    address = "Cambridge, MA",
    publisher = "MIT Press",
    pages = "559--601"
}

@inproceedings{daza-frank-2020-x,
    title = "{X}-{SRL}: A Parallel Cross-Lingual Semantic Role Labeling Dataset",
    author = "Daza, Angel  and
      Frank, Anette",
    editor = "Webber, Bonnie  and
      Cohn, Trevor  and
      He, Yulan  and
      Liu, Yang",
    booktitle = "Proceedings of the 2020 Conference on Empirical Methods in Natural Language Processing (EMNLP)",
    month = nov,
    year = "2020",
    address = "Online",
    publisher = "Association for Computational Linguistics",
    pages = "3904--3914",
}

@inproceedings{lauscher-etal-2020-zero,
    title = "From Zero to Hero: {O}n the Limitations of Zero-Shot Language Transfer with Multilingual {T}ransformers",
    author = "Lauscher, Anne  and
      Ravishankar, Vinit  and
      Vuli{\'c}, Ivan  and
      Glava{\v{s}}, Goran",
    booktitle = "Proceedings of the 2020 Conference on Empirical Methods in Natural Language Processing (EMNLP)",
    month = nov,
    year = "2020",
    address = "Online",
    publisher = "Association for Computational Linguistics",
    pages = "4483--4499",
}

@article{artetxe-schwenk-2019-massively,
    title = "Massively Multilingual Sentence Embeddings for Zero-Shot Cross-Lingual Transfer and Beyond",
    author = "Artetxe, Mikel  and
      Schwenk, Holger",
    journal = "Transactions of the Association for Computational Linguistics",
    volume = "7",
    year = "2019",
    address = "Cambridge, MA",
    publisher = "MIT Press",
    pages = "597--610",
}

@inproceedings{peters-etal-2019-tune,
    title = "To Tune or Not to Tune? Adapting Pretrained Representations to Diverse Tasks",
    author = "Peters, Matthew E.  and
      Ruder, Sebastian  and
      Smith, Noah A.",
    booktitle = "Proceedings of the 4th Workshop on Representation Learning for NLP (RepL4NLP-2019)",
    month = aug,
    year = "2019",
    address = "Florence, Italy",
    publisher = "Association for Computational Linguistics",
    pages = "7--14",
}

@article{Palmer2005,
author = {Palmer, Martha and Gildea, Daniel and Kingsbury, Paul},
title = {{The Proposition Bank: An Annotated Corpus of Semantic Roles}},
year = {2005},
issue_date = {March 2005},
publisher = {MIT Press},
address = {Cambridge, MA, USA},
volume = {31},
number = {1},
issn = {0891-2017},
journal = {Computational Linguistics},
month = {March},
pages = {71--106},
numpages = {36}
}

@inproceedings{arviv-etal-2021-relation,
    title = "On the Relation between Syntactic Divergence and Zero-Shot Performance",
    author = "Arviv, Ofir  and
      Nikolaev, Dmitry  and
      Karidi, Taelin  and
      Abend, Omri",
    booktitle = "Proceedings of the 2021 Conference on Empirical Methods in Natural Language Processing",
    month = nov,
    year = "2021",
    address = "Online and Punta Cana, Dominican Republic",
    publisher = "Association for Computational Linguistics",
    pages = "4803--4817",
}

@inproceedings{briakou-carpuat-2020-detecting,
    title = "{D}etecting {F}ine-{G}rained {C}ross-{L}ingual {S}emantic {D}ivergences without {S}upervision by {L}earning to {R}ank",
    author = "Briakou, Eleftheria  and
      Carpuat, Marine",
    booktitle = "Proceedings of the 2020 Conference on Empirical Methods in Natural Language Processing (EMNLP)",
    month = nov,
    year = "2020",
    address = "Online",
    publisher = "Association for Computational Linguistics",
    pages = "1563--1580",
}

@article{edmonds-hirst-2002-near,
    title = "Near-Synonymy and Lexical Choice",
    author = "Edmonds, Philip  and
      Hirst, Graeme",
    journal = "Computational Linguistics",
    volume = "28",
    number = "2",
    year = "2002",
    address = "Cambridge, MA",
    publisher = "MIT Press",
    pages = "105--144",
}

@inproceedings{schwenk-etal-2021-wikimatrix,
    title = "{W}iki{M}atrix: Mining 135{M} Parallel Sentences in 1620 Language Pairs from {W}ikipedia",
    author = "Schwenk, Holger  and
      Chaudhary, Vishrav  and
      Sun, Shuo  and
      Gong, Hongyu  and
      Guzm{\'a}n, Francisco",
    booktitle = "Proceedings of the 16th Conference of the European Chapter of the Association for Computational Linguistics: Main Volume",
    month = apr,
    year = "2021",
    address = "Online",
    publisher = "Association for Computational Linguistics",
    pages = "1351--1361",
}

@inproceedings{pires-etal-2019-multilingual,
    title = "How Multilingual is Multilingual {BERT}?",
    author = "Pires, Telmo  and
      Schlinger, Eva  and
      Garrette, Dan",
    booktitle = "Proceedings of the 57th Annual Meeting of the Association for Computational Linguistics",
    month = jul,
    year = "2019",
    address = "Florence, Italy",
    publisher = "Association for Computational Linguistics",
    pages = "4996--5001",
}

@inproceedings{keung-etal-2020-dont,
    title = "Don{'}t Use {E}nglish Dev: On the Zero-Shot Cross-Lingual Evaluation of Contextual Embeddings",
    author = "Keung, Phillip  and
      Lu, Yichao  and
      Salazar, Julian  and
      Bhardwaj, Vikas",
    booktitle = "Proceedings of the 2020 Conference on Empirical Methods in Natural Language Processing (EMNLP)",
    month = nov,
    year = "2020",
    address = "Online",
    publisher = "Association for Computational Linguistics",
    pages = "549--554",
}

@inproceedings{abend-rappoport-2013-universal,
    title = "{U}niversal {C}onceptual {C}ognitive {A}nnotation ({UCCA})",
    author = "Abend, Omri  and
      Rappoport, Ari",
    booktitle = "Proceedings of the 51st Annual Meeting of the Association for Computational Linguistics (Volume 1: Long Papers)",
    month = aug,
    year = "2013",
    address = "Sofia, Bulgaria",
    publisher = "Association for Computational Linguistics",
    pages = "228--238",
}

@inproceedings{Banarescu2013,
    title = {{A}bstract {M}eaning {R}epresentation for Sembanking"},
    author = {Laura Banarescu  and
      Claire Bonial  and
      Shu Cai  and
      Madalina Georgescu  and
      Kira Griffitt  and
      Ulf Hermjakob  and
      Kevin Knight and
      Philipp Koehn  and
      Martha Palmer  and
      Nathan Schneider},
    booktitle = {{Proceedings of the 7th Linguistic Annotation Workshop and Interoperability with Discourse}},
    month = {August},
    year = {2013},
    address = {Sofia, Bulgaria},
    publisher = {Association for Computational Linguistics},
    pages = {178--186}
}

@inproceedings{sulem-etal-2015-conceptual,
    title = "Conceptual Annotations Preserve Structure Across Translations: A {F}rench-{E}nglish Case Study",
    author = "Sulem, Elior  and
      Abend, Omri  and
      Rappoport, Ari",
    booktitle = "Proceedings of the 1st Workshop on Semantics-Driven Statistical Machine Translation ({S}2{MT} 2015)",
    month = jul,
    year = "2015",
    address = "Beijing, China",
    publisher = "Association for Computational Linguistics",
    pages = "11--22",
}

@inproceedings{hajic2003pdt,
  title={{PDT-VALLEX}: {C}reating a large-coverage valency lexicon for treebank annotation},
  author={Hajic, Jan and Panevov{\'a}, Jarmila and Ure{\v{s}}ov{\'a}, Zdenka and B{\'e}mov{\'a}, Alevtina and Kol{\'a}rov{\'a}, Veronika and Pajas, Petr},
  booktitle={Proceedings of the second workshop on treebanks and linguistic theories},
  volume={9},
  pages={57--68},
  year={2003}
}

@inproceedings{hajic-etal-2012-announcing,
    title = "Announcing {P}rague {C}zech-{E}nglish {D}ependency {T}reebank 2.0",
    author = "Haji{\v{c}}, Jan  and
      Haji{\v{c}}ov{\'a}, Eva  and
      Panevov{\'a}, Jarmila  and
      Sgall, Petr  and
      Bojar, Ond{\v{r}}ej  and
      Cinkov{\'a}, Silvie  and
      Fu{\v{c}}{\'\i}kov{\'a}, Eva  and
      Mikulov{\'a}, Marie  and
      Pajas, Petr  and
      Popelka, Jan  and
      Semeck{\'y}, Ji{\v{r}}{\'\i}  and
      {\v{S}}indlerov{\'a}, Jana  and
      {\v{S}}t{\v{e}}p{\'a}nek, Jan  and
      Toman, Josef  and
      Ure{\v{s}}ov{\'a}, Zde{\v{n}}ka  and
      {\v{Z}}abokrtsk{\'y}, Zden{\v{e}}k",
    booktitle = "Proceedings of the Eighth International Conference on Language Resources and Evaluation ({LREC}'12)",
    month = may,
    year = "2012",
    address = "Istanbul, Turkey",
    publisher = "European Language Resources Association (ELRA)",
    pages = "3153--3160",
}

@inproceedings{vsindlerova2013verb,
  title={Verb valency and argument noncorrespondence in a bilingual treebank},
  author={{\v{S}}indlerov{\'a}, Jana and Ure{\v{s}}ov{\'a}, Zde{\v{n}}ka and Fu{\v{c}}{\'\i}kov{\'a}, Eva},
  booktitle={Proceedings of the Seventh International Conference Slovko},
  pages={100--108},
  year={2013}
}

@inproceedings{nivre-etal-2016-universal,
    title = "{U}niversal {D}ependencies v1: A Multilingual Treebank Collection",
    author = "Nivre, Joakim  and
      de Marneffe, Marie-Catherine  and
      Ginter, Filip  and
      Goldberg, Yoav  and
      Haji{\v{c}}, Jan  and
      Manning, Christopher D.  and
      McDonald, Ryan  and
      Petrov, Slav  and
      Pyysalo, Sampo  and
      Silveira, Natalia  and
      Tsarfaty, Reut  and
      Zeman, Daniel",
    booktitle = "Proceedings of the Tenth International Conference on Language Resources and Evaluation ({LREC}'16)",
    month = may,
    year = "2016",
    address = "Portoro{\v{z}}, Slovenia",
    publisher = "European Language Resources Association (ELRA)",
    pages = "1659--1666",
}

@inproceedings{nikolaev-etal-2020-fine,
    title = "Fine-Grained Analysis of Cross-Linguistic Syntactic Divergences",
    author = "Nikolaev, Dmitry  and
      Arviv, Ofir  and
      Karidi, Taelin  and
      Kenneth, Neta  and
      Mitnik, Veronika  and
      Saeboe, Lilja Maria  and
      Abend, Omri",
    booktitle = "Proceedings of the 58th Annual Meeting of the Association for Computational Linguistics",
    month = jul,
    year = "2020",
    address = "Online",
    publisher = "Association for Computational Linguistics",
    pages = "1159--1176",
}

@inproceedings{ni-etal-2017-weakly,
    title = "Weakly Supervised Cross-Lingual Named Entity Recognition via Effective Annotation and Representation Projection",
    author = "Ni, Jian  and
      Dinu, Georgiana  and
      Florian, Radu",
    booktitle = "Proceedings of the 55th Annual Meeting of the Association for Computational Linguistics (Volume 1: Long Papers)",
    month = jul,
    year = "2017",
    address = "Vancouver, Canada",
    publisher = "Association for Computational Linguistics",
    pages = "1470--1480",
}

@inproceedings{johannsen-etal-2016-joint,
    title = "Joint part-of-speech and dependency projection from multiple sources",
    author = "Johannsen, Anders  and
      Agi{\'c}, {\v{Z}}eljko  and
      S{\o}gaard, Anders",
    booktitle = "Proceedings of the 54th Annual Meeting of the Association for Computational Linguistics (Volume 2: Short Papers)",
    month = aug,
    year = "2016",
    address = "Berlin, Germany",
    publisher = "Association for Computational Linguistics",
    pages = "561--566",
}

@inproceedings{vyas-etal-2018-identifying,
    title = "Identifying Semantic Divergences in Parallel Text without Annotations",
    author = "Vyas, Yogarshi  and
      Niu, Xing  and
      Carpuat, Marine",
    booktitle = "Proceedings of the 2018 Conference of the North {A}merican Chapter of the Association for Computational Linguistics: Human Language Technologies, Volume 1 (Long Papers)",
    month = jun,
    year = "2018",
    address = "New Orleans, Louisiana",
    publisher = "Association for Computational Linguistics",
    pages = "1503--1515",
}

@inproceedings{wein-etal-2022-effect,
    title = "Effect of Source Language on {AMR} Structure",
    author = "Wein, Shira  and
      Leung, Wai Ching  and
      Mu, Yifu  and
      Schneider, Nathan",
    booktitle = "Proceedings of the 16th Linguistic Annotation Workshop (LAW-XVI) within LREC2022",
    month = jun,
    year = "2022",
    address = "Marseille, France",
    publisher = "European Language Resources Association",
    pages = "97--102",
}

@inproceedings{xue-etal-2014-interlingua,
    title = "Not an Interlingua, But Close: Comparison of {E}nglish {AMR}s to {C}hinese and {C}zech",
    author = "Xue, Nianwen  and
      Bojar, Ond{\v{r}}ej  and
      Haji{\v{c}}, Jan  and
      Palmer, Martha  and
      Ure{\v{s}}ov{\'a}, Zde{\v{n}}ka  and
      Zhang, Xiuhong",
    booktitle = "Proceedings of the Ninth International Conference on Language Resources and Evaluation ({LREC}'14)",
    month = may,
    year = "2014",
    address = "Reykjavik, Iceland",
    publisher = "European Language Resources Association (ELRA)",
    pages = "1765--1772",
}

@inproceedings{ormazabal-etal-2019-analyzing,
    title = "Analyzing the Limitations of Cross-lingual Word Embedding Mappings",
    author = "Ormazabal, Aitor  and
      Artetxe, Mikel  and
      Labaka, Gorka  and
      Soroa, Aitor  and
      Agirre, Eneko",
    booktitle = "Proceedings of the 57th Annual Meeting of the Association for Computational Linguistics",
    month = jul,
    year = "2019",
    address = "Florence, Italy",
    publisher = "Association for Computational Linguistics",
    pages = "4990--4995",
}

@article{DBLP:journals/ipm/EronenPM23,
  author       = {Juuso Eronen and
                  Michal Ptaszynski and
                  Fumito Masui},
  title        = {Zero-shot cross-lingual transfer language selection using linguistic
                  similarity},
  journal      = {Inf. Process. Manag.},
  volume       = {60},
  number       = {3},
  pages        = {103250},
  year         = {2023},
}

@article{agic-etal-2016-multilingual,
    title = "Multilingual Projection for Parsing Truly Low-Resource Languages",
    author = "Agi{\'c}, {\v{Z}}eljko  and
      Johannsen, Anders  and
      Plank, Barbara  and
      Mart{\'\i}nez Alonso, H{\'e}ctor  and
      Schluter, Natalie  and
      S{\o}gaard, Anders",
    journal = "Transactions of the Association for Computational Linguistics",
    volume = "4",
    year = "2016",
    address = "Cambridge, MA",
    publisher = "MIT Press",
    pages = "301--312",
}

@inproceedings{mcdonald-etal-2013-universal,
    title = "{U}niversal {D}ependency Annotation for Multilingual Parsing",
    author = {McDonald, Ryan  and
      Nivre, Joakim  and
      Quirmbach-Brundage, Yvonne  and
      Goldberg, Yoav  and
      Das, Dipanjan  and
      Ganchev, Kuzman  and
      Hall, Keith  and
      Petrov, Slav  and
      Zhang, Hao  and
      T{\"a}ckstr{\"o}m, Oscar  and
      Bedini, Claudia  and
      Bertomeu Castell{\'o}, N{\'u}ria  and
      Lee, Jungmee},
    booktitle = "Proceedings of the 51st Annual Meeting of the Association for Computational Linguistics (Volume 2: Short Papers)",
    month = aug,
    year = "2013",
    address = "Sofia, Bulgaria",
    publisher = "Association for Computational Linguistics",
    pages = "92--97",
}

@inproceedings{blloshmi-etal-2020-xl,
    title = "{XL}-{AMR}: Enabling Cross-Lingual {AMR} Parsing with Transfer Learning Techniques",
    author = "Blloshmi, Rexhina  and
      Tripodi, Rocco  and
      Navigli, Roberto",
    booktitle = "Proceedings of the 2020 Conference on Empirical Methods in Natural Language Processing (EMNLP)",
    month = nov,
    year = "2020",
    address = "Online",
    publisher = "Association for Computational Linguistics",
    pages = "2487--2500",
}

@inproceedings{wein-schneider-2021-classifying,
    title = "Classifying Divergences in Cross-lingual {AMR} Pairs",
    author = "Wein, Shira  and
      Schneider, Nathan",
    booktitle = "Proceedings of the Joint 15th Linguistic Annotation Workshop (LAW) and 3rd Designing Meaning Representations (DMR) Workshop",
    month = nov,
    year = "2021",
    address = "Punta Cana, Dominican Republic",
    publisher = "Association for Computational Linguistics",
    pages = "56--65",
}

@inproceedings{carpuat-etal-2017-detecting,
    title = "Detecting Cross-Lingual Semantic Divergence for Neural Machine Translation",
    author = "Carpuat, Marine  and
      Vyas, Yogarshi  and
      Niu, Xing",
    booktitle = "Proceedings of the First Workshop on Neural Machine Translation",
    month = aug,
    year = "2017",
    address = "Vancouver",
    publisher = "Association for Computational Linguistics",
    pages = "69--79",
}

@inproceedings{ijcai2021p539,
  title     = {MultiMirror: Neural Cross-lingual Word Alignment for Multilingual Word Sense Disambiguation},
  author    = {Procopio, Luigi and Barba, Edoardo and Martelli, Federico and Navigli, Roberto},
  booktitle = {Proceedings of the Thirtieth International Joint Conference on
               Artificial Intelligence, {IJCAI-21}},
  publisher = {International Joint Conferences on Artificial Intelligence Organization},
  editor    = {Zhi-Hua Zhou},
  pages     = {3915--3921},
  year      = {2021},
  month     = {8},
  note      = {Main Track},
}

@inproceedings{jalili-sabet-etal-2020-simalign,
    title = "{S}im{A}lign: High Quality Word Alignments Without Parallel Training Data Using Static and Contextualized Embeddings",
    author = {Jalili Sabet, Masoud  and
      Dufter, Philipp  and
      Yvon, Fran{\c{c}}ois  and
      Sch{\"u}tze, Hinrich},
    booktitle = "Findings of the Association for Computational Linguistics: EMNLP 2020",
    month = nov,
    year = "2020",
    address = "Online",
    publisher = "Association for Computational Linguistics",
    pages = "1627--1643",
}

@inproceedings{baker-etal-1998-berkeley-framenet,
    title = "The {B}erkeley {F}rame{N}et Project",
    author = "Baker, Collin F.  and
      Fillmore, Charles J.  and
      Lowe, John B.",
    booktitle = "36th Annual Meeting of the Association for Computational Linguistics and 17th International Conference on Computational Linguistics, Volume 1",
    month = aug,
    year = "1998",
    address = "Montreal, Quebec, Canada",
    publisher = "Association for Computational Linguistics",
    pages = "86--90",
}

@inproceedings{miller-1992-wordnet,
    title = "{W}ord{N}et: A Lexical Database for {E}nglish",
    author = "Miller, George A.",
    booktitle = "Speech and Natural Language: Proceedings of a Workshop Held at Harriman, New York, {F}ebruary 23-26, 1992",
    year = "1992",
}

@article{li1990xiandai,
  title={Xiandai {H}anyu {D}ongci ({M}odern {C}hinese {V}erbs)},
  author={Li, LD},
  journal={China Social Science Press, Beijing},
  year={1990}
}

@inproceedings{conneau-etal-2020-unsupervised,
    title = "Unsupervised Cross-lingual Representation Learning at Scale",
    author = "Conneau, Alexis  and
      Khandelwal, Kartikay  and
      Goyal, Naman  and
      Chaudhary, Vishrav  and
      Wenzek, Guillaume  and
      Guzm{\'a}n, Francisco  and
      Grave, Edouard  and
      Ott, Myle  and
      Zettlemoyer, Luke  and
      Stoyanov, Veselin",
    booktitle = "Proceedings of the 58th Annual Meeting of the Association for Computational Linguistics",
    month = jul,
    year = "2020",
    address = "Online",
    publisher = "Association for Computational Linguistics",
    pages = "8440--8451",
}

@inproceedings{devlin-etal-2019-bert,
    title = "{BERT}: Pre-training of Deep Bidirectional Transformers for Language Understanding",
    author = "Devlin, Jacob  and
      Chang, Ming-Wei  and
      Lee, Kenton  and
      Toutanova, Kristina",
    booktitle = "Proceedings of the 2019 Conference of the North {A}merican Chapter of the Association for Computational Linguistics: Human Language Technologies, Volume 1 (Long and Short Papers)",
    month = jun,
    year = "2019",
    address = "Minneapolis, Minnesota",
    publisher = "Association for Computational Linguistics",
    pages = "4171--4186",
}

@article{ponti-etal-2021-parameter,
    title = "Parameter Space Factorization for Zero-Shot Learning across Tasks and Languages",
    author = "Ponti, Edoardo M.  and
      Vuli{\'c}, Ivan  and
      Cotterell, Ryan  and
      Parovic, Marinela  and
      Reichart, Roi  and
      Korhonen, Anna",
    journal = "Transactions of the Association for Computational Linguistics",
    volume = "9",
    year = "2021",
    address = "Cambridge, MA",
    publisher = "MIT Press",
    pages = "410--428",
}

@article{navigli2012babelnet,
  title={BabelNet: The automatic construction, evaluation and application of a wide-coverage multilingual semantic network},
  author={Navigli, Roberto and Ponzetto, Simone Paolo},
  journal={Artificial intelligence},
  volume={193},
  pages={217--250},
  year={2012},
  publisher={Elsevier}
}

@inproceedings{agic-etal-2014-cross,
    title = "Cross-lingual Dependency Parsing of Related Languages with Rich Morphosyntactic Tagsets",
    author = {Agi{\'c}, {\v{Z}}eljko  and
      Tiedemann, J{\"o}rg  and
      Merkler, Danijela  and
      Krek, Simon  and
      Dobrovoljc, Kaja  and
      Mo{\v{z}}e, Sara},
    booktitle = "Proceedings of the {EMNLP}{'}2014 Workshop on Language Technology for Closely Related Languages and Language Variants",
    month = "oct",
    year = "2014",
    address = "Doha, Qatar",
    publisher = "Association for Computational Linguistics",
    pages = "13--24"
}

@inproceedings{gerz-etal-2018-relation,
    title = "On the Relation between Linguistic Typology and (Limitations of) Multilingual Language Modeling",
    author = "Gerz, Daniela  and
      Vuli{\'c}, Ivan  and
      Ponti, Edoardo Maria  and
      Reichart, Roi  and
      Korhonen, Anna",
    booktitle = "Proceedings of the 2018 Conference on Empirical Methods in Natural Language Processing",
    month = oct # "-" # nov,
    year = "2018",
    address = "Brussels, Belgium",
    publisher = "Association for Computational Linguistics",
    pages = "316--327",
}

@inproceedings{ruder-etal-2019-transfer,
    title = "Transfer Learning in Natural Language Processing",
    author = "Ruder, Sebastian  and
      Peters, Matthew E.  and
      Swayamdipta, Swabha  and
      Wolf, Thomas",
    booktitle = "Proceedings of the 2019 Conference of the North {A}merican Chapter of the Association for Computational Linguistics: Tutorials",
    month = jun,
    year = "2019",
    address = "Minneapolis, Minnesota",
    publisher = "Association for Computational Linguistics",
    pages = "15--18"
}

@inproceedings{DBLP:conf/nips/ConneauL19,
  author       = {Alexis Conneau and
                  Guillaume Lample},
  editor       = {Hanna M. Wallach and
                  Hugo Larochelle and
                  Alina Beygelzimer and
                  Florence d'Alch{\'{e}}{-}Buc and
                  Emily B. Fox and
                  Roman Garnett},
  title        = {Cross-lingual Language Model Pretraining},
  booktitle    = {Advances in Neural Information Processing Systems 32: Annual Conference
                  on Neural Information Processing Systems 2019, NeurIPS 2019, December
                  8-14, 2019, Vancouver, BC, Canada},
  pages        = {7057--7067},
  year         = {2019}
}

@inproceedings{conneau-etal-2018-xnli,
    title = "{XNLI}: Evaluating Cross-lingual Sentence Representations",
    author = "Conneau, Alexis  and
      Rinott, Ruty  and
      Lample, Guillaume  and
      Williams, Adina  and
      Bowman, Samuel  and
      Schwenk, Holger  and
      Stoyanov, Veselin",
    booktitle = "Proceedings of the 2018 Conference on Empirical Methods in Natural Language Processing",
    month = oct # "-" # nov,
    year = "2018",
    address = "Brussels, Belgium",
    publisher = "Association for Computational Linguistics",
    pages = "2475--2485"
}

@inproceedings{di2019verbatlas,
  title={VerbAtlas: a novel large-scale verbal semantic resource and its application to semantic role labeling},
  author={Di Fabio, Andrea and Conia, Simone and Navigli, Roberto},
  booktitle={Proceedings of the 2019 Conference on Empirical Methods in Natural Language Processing and the 9th International Joint Conference on Natural Language Processing (EMNLP-IJCNLP)},
  pages={627--637},
  year={2019}
}

@article{pado2009cross,
  title={Cross-lingual annotation projection for semantic roles},
  author={Pad{\'o}, Sebastian and Lapata, Mirella},
  journal={Journal of Artificial Intelligence Research},
  volume={36},
  pages={307--340},
  year={2009}
}

@inproceedings{david2001inducing,
  title={Inducing multilingual text analysis tools via robust projection across aligned corpora},
  author={Yarowsky, David and Grace, Ngai and Richard, Wicentowski and others},
  booktitle={Proceedings of the First International Conference on Human Language Technology Research},
  pages={1--8},
  year={2001}
}

@article{dorr1994machine,
  title={Machine translation divergences: A formal description and proposed solution},
  author={Dorr, Bonnie},
  journal={Computational linguistics},
  volume={20},
  number={4},
  pages={597--633},
  year={1994}
}

@inproceedings{tripodi2021united,
  title={UniteD-SRL: A unified dataset for span-and dependency-based multilingual and cross-lingual Semantic Role Labeling},
  author={Tripodi, Rocco and Conia, Simone and Navigli, Roberto},
  booktitle={Findings of the Association for Computational Linguistics: EMNLP 2021},
  pages={2293--2305},
  year={2021}
}

\clearpage
\newpage
\newpage

\appendix\section{APPENDIX}\label{sec:appendix}

\paragraph{List of VerbAtlas Frames used in this Work with definitions}

\begin{longtable}{p{5cm}p{8cm}}\label{tab:frame_definitions}\\
\caption{Definition of the VerbAtlas frames used in this work. The description of the arguments structure of the frame is shown when the definition of a given frame is missing in VerbAtlas.}\\
\toprule
\textsc{\lowercase{AFFIRM}} & To declare or affirm solemnly and formally as true\\
\textsc{\lowercase{REQUIRE\_NEED\_WANT\_HOPE}} & A cause makes an agent REQUIRE-NEED-WANT-HOPE for a theme from a source for a beneficiary to achieve a goal in exchange for a co-theme\\
\textsc{\lowercase{CONTINUE}} & An agent with a co-agent make a theme CONTINUE from a source to a destination using an instrument (+attribute)\\
\textsc{\lowercase{DISCUSS}} & An agent \textsc{\lowercase{DISCUSSES}} with a co-agent about a topic (+attribute)\\
\textsc{\lowercase{SEND}} & An agent \textsc{\lowercase{SENDS}} a theme to a destination\\
\textsc{\lowercase{AUTHORIZE\_ADMIT}} & Admit into a group or community\\
\textsc{\lowercase{NAME}} & Mark or expose as infamous\\
\textsc{\lowercase{SPEED-UP}} & Move faster\\
\textsc{\lowercase{ALLY\_ASSOCIATE\_MARRY}} & Give to in marriage\\
\textsc{\lowercase{PRECLUDE\_FORBID\_EXPEL}} & Hinder or prevent (the efforts, plans, or desires) of\\
\textsc{\lowercase{USE}} & Put into service; make work or employ for a particular purpose or for its inherent or natural purpose\\
\textsc{\lowercase{RETAIN\_KEEP\_SAVE-MONEY}} & Be designed to hold or take\\
\textsc{\lowercase{MOUNT\_ASSEMBLE\_PRODUCE}} & Create by putting components or members together\\
\textsc{\lowercase{CHANGE-HANDS}} & Transfer property or ownership\\
\textsc{\lowercase{DESTROY}} & An agent \textsc{\lowercase{DESTROYS}} a patient with an instrument having a result\\
\textsc{\lowercase{AGREE\_ACCEPT}} & To agree or express agreement\\
\textsc{\lowercase{INCREASE\_ENLARGE\_MULTIPLY}} & An agent \textsc{\lowercase{INCREASES-ENLARGES-MULTIPLIES}} the attribute of a patient of an extent from a source to a destination using an instrument on a location in favour of a beneficiary\\
\textsc{\lowercase{BEGIN}} & Take on titles, offices, duties, responsibilities\\
\textsc{\lowercase{OBLIGE\_FORCE}} & Bind by a contract; especially for a training period\\
\textsc{\lowercase{CARRY-OUT-ACTION}} & Put in effect\\
\textsc{\lowercase{FACE\_CHALLENGE}} & A cause makes an agent \textsc{\lowercase{FACE-CHALLENGE}} a theme with a goal using an instrument (+attribute)\\
\textsc{\lowercase{HAPPEN\_OCCUR}} & Happen, occur, take place\\
\textsc{\lowercase{SIGNAL\_INDICATE}} & An agent \textsc{\lowercase{SIGNALS-INDICATES}} a topic to a recipient using an instrument in a location\\
\textsc{\lowercase{INCLUDE-AS}} & An agent \textsc{\lowercase{INCLUDES}} a patient in a goal with an instrument (+attribute)\\
\textsc{\lowercase{ACHIEVE}} & To gain with effort\\
\textsc{\lowercase{SEE}} & An experiencer \textsc{\lowercase{SEES}} a stimulus in favour of a beneficiary (+attribute)\\
\textsc{\lowercase{INFORM}} & An agent \textsc{\lowercase{INFORMS}} a recipient of a topic using an instrument\\
\textsc{\lowercase{EXEMPT}} & An agent \textsc{\lowercase{EXEMPTS}} a theme from a source\\
\textsc{\lowercase{DRIVE-BACK}} & Expel from a community or group\\
\textsc{\lowercase{EMPHASIZE}} & Put stress on; utter with an accent\\
\textsc{\lowercase{GUARANTEE\_ENSURE\_PROMISE}} & Assure somebody of the truth of something with the intention of giving the listener confidence\\
\textsc{\lowercase{CANCEL\_ELIMINATE}} & Cancel officially\\
\textsc{\lowercase{SPEAK}} & An agent \textsc{\lowercase{SPEAKS}} about a topic in a location to a recipient using an instrument achieving a result (+attribute)\\
\textsc{\lowercase{REDUCE\_DIMINISH}} & An agent \textsc{\lowercase{REDUCES-DIMINISHES}} the attribute of a patient of an extent from a source-state to a goal-state using an instrument on a location\\
\textsc{\lowercase{TAKE}} & Obtain or retrieve from a storage device; as of information on a computer\\
\textsc{\lowercase{SUBJECTIVE-JUDGING}} & Pronounce not guilty of criminal charges\\
\textsc{\lowercase{ESTABLISH}} & Build or establish something abstract\\
\textsc{\lowercase{MANAGE}} & Direct the taking of\\
\textsc{\lowercase{EXIST\_LIVE}} & A theme \textsc{\lowercase{EXISTS-LIVES}} with a co-theme (+attribute)\\
\textsc{\lowercase{PROPOSE}} & Give to understand\\
\textsc{\lowercase{REACH}} & A cause makes a theme \textsc{\lowercase{REACH}} a goal in a location in favour of a beneficiary\\
\textsc{\lowercase{CREATE\_MATERIALIZE}} & Make real or concrete; give reality or substance to\\
\textsc{\lowercase{MATCH}} & Be compatible, similar or consistent; coincide in their characteristics\\
\textsc{\lowercase{MISS\_OMIT\_LACK}} & An agent \textsc{\lowercase{MISSES-OMITS-LACKS}} a theme from a source using an instrument\\
\textsc{\lowercase{EXIST-WITH-FEATURE}} & A cause makes a theme \textsc{\lowercase{EXISTS WITH THE FEATURE}} attribute to achieve a goal having a value\\
\textsc{\lowercase{CHOOSE}} & Vote by ballot\\
\textsc{\lowercase{LEARN}} & Take up mentally\\
\textsc{\lowercase{CIRCULATE\_SPREAD\_DISTRIBUTE}} & An agent \textsc{\lowercase{CIRCULATES-SPREADS-DISTRIBUTES}} a theme to a recipient from a source using an instrument (+attribute)\\
\textsc{\lowercase{PLAN\_SCHEDULE}} & An agent \textsc{\lowercase{PLANS-SCHEDULES}} a theme for a beneficiary at a time to achieve a goal (+attribute)\\
\textsc{\lowercase{PERCEIVE}} & An experiencer \textsc{\lowercase{PERCEIVES}} a stimulus as attribute\\
\textsc{\lowercase{LIBERATE\_ALLOW\_AFFORD}} & Let off the hook\\
\textsc{\lowercase{DISTINGUISH\_DIFFER}} & Be characteristic of\\
\textsc{\lowercase{PROVE}} & Advance evidence for\\
\textsc{\lowercase{INCITE\_INDUCE}} & An agent \textsc{\lowercase{INCITES-INDUCES}} a patient using an instrument to do a result (+attribute)\\
\textsc{\lowercase{REGRET\_SORRY}} & An experiencer \textsc{\lowercase{REGRETS}} is \textsc{\lowercase{SORRY}} for a stimulus to a destination\\
\textsc{\lowercase{COMBINE\_MIX\_UNITE}} & Cause to become one with\\
\bottomrule
\end{longtable}

\paragraph{List of VerbAtlas Roles with Definitions}

\begin{longtable}{p{1.75cm}p{11.5cm}}
\caption{Definition of the VerbAtlas semantic roles.} \label{tab:roles_definitions} \\
\toprule
Agent & Actor in an event who initiates and carries out the event intentionally or consciously, and who exists independently of the event. \\ 
Asset & Value that is a concrete object, usually money.\\
Attribute & Undergoer that is a property of an entity or entities or an entire proposition.\\
Beneficiary & Undergoer in a state or an event that is (potentially) advantaged or disadvantaged by the event or state. \\
Cause & Actor in an event (that may be animate or inanimate) that initiates the event, or cause an agent to initiate an event; it exists independently of the event.\\
Co-Agent & Agent who is acting in coordination or reciprocally with another agent while participating in the same event. \\ 
Co-Patient & Patient that participates in an event with another patient, both participate equally in the event.\\
Co-Theme & Theme that participates in an event or state with another Theme; both participate equally.\\
Destination & Goal of the event that can be a concrete location, a state or a moment in time.\\ 
Experiencer & Patient that is aware of the event undergone (specific to events of perception).\\
Extent & Amount of measurable change to a participant over the course of the event. It can be, for example, an amount of time, distance etc.\\ 
Goal & Abstract place that is the end point of an action and exists independently of the event.\\
Instrument & Undergoer in an event that is manipulated by an agent, and with which an intentional act is performed; it exists independently of the event.\\
Location & Place that is concrete.\\ 
Material & Patient that exists at the starting point of action, which is transformed through the event into a new entity.\\
Patient & Undergoer in an event that experiences a change of state, location or condition, that is causally involved or directly affected by other participants, and exists independently of the event.\\
Product & Result that is a concrete object.\\
Purpose & Purpose of the action.\\
Recipient & Destination that is animate.\\
Result & Goal that comes into existence through the event.\\
Source & Participant that is the starting point of action or event; exists independently of the event.\\
Stimulus & Cause in an event that elicits an emotional or psychological response (specific to events of perception).\\
Time & Participant that indicates an instant or an interval of time during which a state exists or an event took place.\\
Theme & Undergoer that is central to an event or state that does not have control over the way the event occurs, is not structurally changed by the event, and/or is characterized as being in a certain position or condition throughout the state or is something in movement.\\
Topic & Theme characterized by information content.\\ 
Value & Place along a formal scale.\\
\bottomrule
\end{longtable}

\end{document}